\pdfoutput=1

\documentclass[10pt]{article}
\newif\ifblindreview

\usepackage[letterpaper]{geometry}
\usepackage{amta2024}
\usepackage{url}
\usepackage{multicol}
\usepackage{float}
\usepackage[hidelinks]{hyperref}
\usepackage[all]{hypcap}
\hypersetup{
    colorlinks,
    linkcolor={red!50!black},
    citecolor={blue!50!black},
    urlcolor={blue!80!black}
}

\usepackage[T1]{fontenc}
\usepackage[utf8]{inputenc}

\usepackage{times}
\usepackage{latexsym}
\usepackage{booktabs}
\usepackage{natbib}
\usepackage{amsmath,amssymb,amsfonts}
\usepackage{colortbl}
\usepackage{graphicx}
\usepackage{tabularx}
\usepackage[dvipsnames]{xcolor}

\usepackage{listings}
\usepackage{multirow}
\usepackage{rotating}
\usepackage{tablefootnote}
\usepackage{makecell}

\usepackage{enumerate}
\usepackage{lingmacros}
\usepackage{fancyvrb}
\usepackage{tikz}

\usepackage{pifont}
\usepackage{caption}
\usepackage{xspace}

\newcommand{\cmark}{\textcolor{green!70!black}{\ding{51}}}
\newcommand{\xmark}{\textcolor{red}{\ding{55}}}

\usepackage{microtype}
\usepackage{inconsolata}

\definecolor{darkpastelgreen}{rgb}{0.01, 0.75, 0.24}
\definecolor{darkpastelred}{rgb}{0.76, 0.23, 0.13}

\usepackage{xspace}
\xspaceaddexceptions{\$}
\usepackage{placeins}
\usepackage{xurl}

\newcommand{\MTbaseline}{\texttt{baseline}\xspace}
\newcommand{\MTmoslem}{\texttt{base+terms}\xspace}

\newcommand{\ACL}{ParaNLP\xspace}
\newcommand{\iwslt}{IWSLT2023\xspace}

\newcommand{\NV}{\text{NO-VAR}\xspace}
\newcommand{\VG}{\text{GRAPH}\xspace}
\newcommand{\VMS}{\text{MORPH}\xspace}
\newcommand{\VR}{\text{RED}\xspace}
\newcommand{\VE}{\text{EXP}\xspace}
\newcommand{\VL}{\text{LEX}\xspace}
\newcommand{\CM}{\text{COMBO}\xspace}

\newcommand{\Concordancer}{\texorpdfstring{\textsc{Concordancer}}{Concordancer}\xspace}

\newcommand{\ITC}{TCR\xspace}
\newcommand{\ITCnoxspace}{TCR}
\newcommand{\TLC}{TCR\xspace}

\newcommand{\CTV}{CTV\xspace}

\ifblindreview
  \newcommand{\authorinfo}{\author{}}
\else
  \newcommand{\authorinfo}{%
    \author{
        \name{\bf Nicolas Dahan} \hfill \addr{nicolas.dahan@inria.fr}\\
        \addr{\small Inria, Paris, France}
    \AND
        \name{\bf Ziqian Peng} \hfill \addr{peng@isir.upmc.fr}\\
        \addr{\small Sorbonne Université, CNRS, ISIR, Paris, France \& Inria, Paris, France}
    \AND
        \name{\bf François Yvon} \hfill \addr{yvon@isir.upmc.fr}\\
        \addr{\small Sorbonne Université, CNRS, ISIR, Paris, France}
    \AND        
        \name{\bf Rachel Bawden} \hfill
        \addr{rachel.bawden@inria.fr}\\
        \addr{\small Inria, Paris, France}
    }
  }
\fi
\authorinfo

\begin{document}

\amtaHeader{x}{x}{xxx-xxx}{2026}{Term Evaluation in MT: Variation Matters}{Dahan, Peng, Yvon and Bawden}
\title{Improving Term Evaluation in Machine Translation: Variation Matters}
\authorinfo

\maketitle
\pagestyle{empty}

\begin{abstract}
\vspace{5pt}
  Terminology evaluation in machine translation (MT) usually assumes a single correct target form per
  source term. However, human translators routinely introduce variation that current metrics
  penalize as inconsistency. We examine how to account for this variation in document-level MT
  evaluation of English--French scientific translation, combining glossary-based accuracy,
  translation consistency, and a new \emph{cross-term variation} (\CTV) diagnostic measure that tests whether
  variation relationships are preserved across languages. Based on analyses of two parallel corpora, translated by
  four MT systems, we find that (1)~MT systems generate less target-side variation than human translators; (2)~transfer patterns strongly depend on the variation type;
  (3)~consistency rankings vary with the choice of metric; and (4)~constraining MT with a
  glossary improves accuracy and consistency but degrades \CTV\ by suppressing valid
  variation. We argue for variation-aware evaluation that conditions consistency penalties on whether target-side variation mirrors source-side variation.
\end{abstract}

\begin{multicols}{2}

\section{Introduction}
\label{sec:introduction}

Evaluating how machine translation (MT) systems handle terminology in scientific and specialized documents requires accounting for a fundamental property of terms in running texts: \textit{terminological variation}. A  concept (e.g., \textit{machine translation}) does not always appear under a single form, which we call the preferred term. It may surface as an acronym (\textit{MT}), a reduction (\textit{translation}), a lexical substitution (\textit{automatic translation}), or a morphosyntactic reformulation (e.g., \textit{translation by machines})~\citep{daille:hal-01693035}. We refer to these alternative surface forms as variants. Such variation arises naturally in scientific writing, for example because of stylistic alternation, register adaptation, or avoidance of repetition, and is a well-attested feature of specialized discourse \citep{fernandez-silva-kerremans-2011,daille:hal-01693035}.

Whether such variation should be considered an error depends on the role the terminological resource is meant to play. When a glossary prescribes a single sanctioned form, for reasons of trademark, safety, or usability, consistency is genuinely required and any deviation is an error, on both the source and the target side. However, glossaries are just as often provided as vocabulary guidance, and terminology standards themselves recognize a scale of acceptability, from preferred to merely admitted forms \citep{iso1087-2019}. Unless terminology is enforced to be unambiguous and used consistently, as in controlled-language settings, stylistic variation is expected and even preferred, as in the scientific corpora we study. Yet most evaluation metrics penalize such variation uniformly, ignoring whether it mirrors variation already present in the source.

Current MT terminology evaluation does not account for variation. \textit{Glossary-based methods} \citep{alam2021evaluationmachinetranslationterminology,semenov-etal-2023-findings,semenov-etal-2025-findings} match source terms against expected translations from a bilingual glossary, but cannot evaluate non-glossary variants, i.e.~a concept's surface forms that are not listed in the glossary. Such forms naturally arise in running texts, such as reorderings (\textit{quality of translation} instead of \textit{translation quality}), reductions (\textit{processing} for \textit{language processing}), or context-dependent acronyms. Because such forms are produced dynamically as the text unfolds (by insertion, reduction, or new terms being coined from existing ones), they cannot be exhaustively enumerated in advance, and simply enlarging the glossary therefore cannot solve the problem. \textit{Consistency metrics} \citep{itagaki-etal-2007-automatic, lyu-etal-2021-encouraging, semenov-bojar-2022-automated} assess whether a source term is uniformly rendered throughout a document, but considers all target-side variation as inconsistency. Neither family of metrics captures whether the \textit{variation relationship} between a preferred term and its variant, for example between an acronym and its expanded form, is preserved in translation; treating the two as separate terms only measures each in isolation.

In this work, we study how to account for terminological variation in document-level MT evaluation for English--French scientific translation. We extend terminology coverage to non-glossary variants, identified in context by a concordancer (Section~\ref{sec:methodology}), and examine three complementary evaluation dimensions: (1)~term-level accuracy, (2)~translation consistency, both established in prior work, and (3)~cross-term variation, a new dimension introduced to assess whether variation relationships are preserved in translation (see Table~\ref{tab:cross-term-example}). Each dimension has structural coverage gaps tied to variation type, but together they achieve wider coverage. We analyze two parallel English--French scientific corpora from the Natural Language Processing (NLP) domain \citep{salesky-etal-2023-evaluating,peng-etal-2026-parallel}, each translated by four MT systems, and report the following findings:

\begin{itemize}
    \item Human translators introduce more target-side lexical variation than MT systems, reflecting standard translation practices that current consistency metrics penalize.
    \item Variation types are not always preserved in translation, with some types (e.g., reductions) transferred consistently and others (e.g., acronyms) strongly dependent on the corpus and its translation guidelines.
    \item Different measures of translation consistency result in divergent system rankings, showing that existing consistency scores are sensitive to metric design.
    \item Constraining MT to use glossary terms improves accuracy and consistency metrics, but reduces the system's ability to transfer source-side variations onto the target, exposing a tension between terminological uniformity and valid variation.
\end{itemize}

Our contribution is primarily methodological: rather than claim that these variation patterns are universal, we provide a framework and metrics for measuring how terminological variation interacts with MT evaluation, and report empirical findings specific to the two English--French NLP corpora we study.

\begin{table}[H]
\footnotesize
\centering
\setlength{\tabcolsep}{4pt}
\renewcommand{\arraystretch}{0.95}
\begin{tabular}{@{}l p{0.6\columnwidth}@{}}
\toprule
\textbf{Head (EN)} & automatic translation system \\
\textbf{Head (FR)} & système de traduction automatique \\
\textbf{Glossary variants} & none listed \\
\textbf{Src variant} & translation system \textit{(reduction)} \\
\textbf{Expected tgt} & système de traduction \textit{(reduction, inferred)} \\
\textbf{MT output} & système de traduction \textit{(unevaluated by glossary-based methods)} \\
\bottomrule
\end{tabular}
\caption{Cross-term coherence example: the reduction \textit{translation system} is absent from the glossary, but its French counterpart is inferred from the reduction relationship to the head term.}
\label{tab:cross-term-example}
\end{table}

\section{Related Work}
\label{sec:related}

Modern MT evaluation increasingly relies on fine-tuned neural metrics such as COMET \citep{rei-etal-2020-comet} and MetricX \citep{juraska-etal-2024-metricx}, which provide strong overall quality estimates, while LLM-based error analysis approaches such as GEMBA-MQM \citep{kocmi-federmann-2023-gemba} produce fine-grained Multidimensional Quality Metrics (MQM) annotations that include terminology errors. These methods operate at the segment level, however, and do not model term-level structure: they neither track per-concept cross-document consistency nor capture variation relationships between preferred terms and variants. This holds most directly for the neural metrics, which score semantic adequacy rather than surface form: they rarely penalize valid variation but, for the same reason, cannot diagnose terminological inconsistency. GEMBA-MQM goes beyond adequacy, targeting terminology through a dedicated MQM category, but its predicted error spans do not align well with human annotations \citep{lu-etal-2025-mqm}, and, scoring each segment in isolation, it does not track whether a concept is rendered consistently across a document or whether variation relationships are preserved. Dedicated term-level evaluation protocols have therefore been developed to close these gaps.

A first family of term-level methods relies on a bilingual glossary, measuring how often MT produces the expected target term for each source entry. Matching strategies range from exact surface-form matching \citep{alam2021evaluationmachinetranslationterminology} to partial and lemma-based matching \citep{alam-etal-2021-findings, semenov-etal-2025-findings}, and have been adopted in shared evaluation campaigns such as the WMT terminology translation tasks \citep{alam-etal-2021-findings, semenov-etal-2023-findings, semenov-etal-2025-findings}. Document-level extensions further compute per-term ratios of target-to-source occurrences \citep{semenov-etal-2025-findings}. Even with lemmatization, these approaches remain centered on canonical glossary entries and miss unlisted surface forms such as reductions, dynamic acronyms, and lexical variants. 

Glossaries are also useful to enforce specific translations at generation time through lexically constrained decoding \citep{hokamp-liu-2017-lexically, post-vilar-2018-fast} or constraint-aware training \citep{dinu-etal-2019-training}, with related lexical constraints used in paraphrase generation \citep{hu-etal-2019-parabank}; our glossary-guided condition (Section~\ref{sec:setup}) instead applies such guidance through prompting rather than hard decoding constraints.

When no glossary is available, consistency-based metrics offer an alternative. \citet{semenov-bojar-2022-automated} compute per-term consistency with respect to a pseudo-reference derived from the first target realization. Concentration-based metrics, such as the Herfindahl--Hirschman Index \citep{itagaki-etal-2007-automatic, Gapar2022MeasuringTC} and the Lexical Translation Consistency Ratio (LTCR) \citep{lyu-etal-2021-encouraging, wang-etal-2025-delta} quantify how uniformly a source term is translated, by rewarding the reuse of a dominant form. Despite methodological differences, these methods share a common assumption: optimal scores require that every occurrence of a term should receive the same translation. They therefore penalize acceptable target-side variation, such as the stylistic alternation commonly observed in scientific writing \citep{vinay-darbelnet-1995}, where alternating between forms is a deliberate, well-motivated device rather than carelessness \citep{bowker-1998-variant, pecman-2014-variation}, and enforcing a single form risks over-standardization that erases author-intended distinctions \citep{bowker-hawkins-2006-variation}.

This assumption sits at odds with a long line of linguistic work showing that terminological variation is a fundamental property of specialized discourse rather than random noise. \citet{daille:hal-01693035} provides a comprehensive characterization of denominative variation in domain-specific corpora, distinguishing graphical, morphosyntactic, semantic, and pragmatic variants of a single concept; \citet{carcamo:hal-05442584} refine this typology into graphical, morphosyntactic, reduction, expansion, and lexical variants. Translation studies have also started to examine how variation behaves under translation: \citet{fernandez-silva-kerremans-2011} show that source-side variation is broadly preserved or accentuated in human translations of Galician scientific texts into English, while \citet{culo-nitzke-2016-patterns} observe, in English--German post-edited texts, that MT systems produce fewer variant translations than humans, with MT patterns carrying over into post-editing. This reduction is consistent with broader evidence that MT produces lexically less diverse and less rich output than human translators \citep{vanmassenhove-etal-2019-lost,vanmassenhove-etal-2021-machine}, and with the notion of \textit{translationese}, of which post-edited MT is an exacerbated case: it is simpler and more normalized than human translation \citep{toral-2019-post}. These descriptive findings have not yet been integrated into automatic evaluation: none of these metrics model how variation types affect evaluation coverage, nor whether structural relationships between preferred terms and variants (e.g., acronym--expansion pairs) are preserved in translation.

\section{Methodology}
\label{sec:methodology}

We study how MT systems translate terms at the document level, accounting for the range of surface forms a term may assume in context, along three complementary dimensions: (1)~how accurately each occurrence is translated based on a reference glossary, (2)~how consistently each surface form (whether a preferred term or a variant) is rendered throughout its document occurrences, and (3)~whether variation relationships between terms and variants are preserved in translation.

Our terminological resource is structured as a Simple Knowledge Organization System (SKOS)\footnote{\url{https://www.w3.org/TR/skos-reference/}} glossary: each entry pairs a concept with a preferred term (SKOS \texttt{prefLabel}) and a list of alternative terms (SKOS \texttt{altLabel}); we use \textit{variant} as a synonym for alternative term. We turn term occurrences into their canonical forms by stripping inflection and other non-terminological variation (e.g., by lemmatization and orthographic normalization). Our pipeline takes this glossary as input and processes parallel documents in four steps: (1)~term detection, (2)~source-side variation labeling, (3)~source-target alignment, and (4)~target-side labeling. The output is a list of annotated term pairs relating each source occurrence with the corresponding target span.

\begin{table}[H]
\centering\small
\setlength{\tabcolsep}{4pt}
\renewcommand{\arraystretch}{0.92}
\begin{tabular}{@{}lp{0.78\columnwidth}@{}}
\toprule
\textbf{Type} & \textbf{Description and example} \\
\midrule
\VG  & Graphical (acronyms, spelling): \textit{intellectual property / IP} \\
\VMS & Morphosyntactic (reordering, inflection): \textit{contract of employment / employment contract} \\
\VR  & Reduction (truncation): \textit{European Union acquis / acquis} \\
\VE  & Expansion (added material): \textit{wear / normal wear and tear} \\
\VL  & Lexical (near-synonymy): \textit{action for damages / claim for damages} \\
\CM  & Combined (multiple types): \textit{coronavirus disease 2019 / COVID-19} \\
\bottomrule
\end{tabular}
\caption{Typology used to label variation in source- and target term occurrences \citep{carcamo:hal-05442584}.}
\label{tab:variation-types-summary}
\end{table}

\paragraph{Term detection.}
We identify occurrences of known terms using \Concordancer,\footnote{\url{https://gitlab.inria.fr/almanach/concordancer}} a trie-based concordancer
operating on documents to which linguistic analysis (lemmatization, POS tagging, dependency parsing) is applied with spaCy \citep{spacy}. Coverage extends beyond exact glossary entries via two complementary mechanisms: (1)~\textit{static variants}, generated offline from each glossary entry by applying a fixed set of transformation rules (synonym substitution via WordNet \citep{miller-1995-wordnet}, syntactic restructuring, morphological derivation via MorphyNet \citep{batsuren-etal-2021-morphynet}, and modifier reduction); (2)~\textit{dynamic variants}, detected at run-time from dependency patterns in context. Together these recover a wider range of forms under which each concept appears in the document.

\paragraph{Source-side variation labeling.}
Each detected occurrence is labeled with its variation type relative to the preferred term, following the typology of \citet{carcamo:hal-05442584} (Table~\ref{tab:variation-types-summary}; full typology in Appendix~\ref{app:variation-typology}). The label provenance depends on whether the matched form is lexicalized in the glossary: for the preferred term and its alternative terms, we assign the type by few-shot prompting ChatGPT (\texttt{gpt-4.1-mini}; prompts in Appendix~\ref{app:variation-classification-prompt}); for non-lexicalized matches (static or dynamic variants), the type is derived directly from the \Concordancer\ formation pattern that triggered the match. When a non-lexicalized match is derived from an alternative term rather than from the preferred term, the final label combines both classifications (e.g., a dynamic acronym of an \VE\ variant becomes \CM).
To validate this step, one of the authors, a computational linguist and native French speaker fluent in English, manually assigned gold variation labels to 50~sampled pairs; the classifier agrees with these human labels on 84\% of types and 77\% of subtypes, with near-perfect agreement on single-type variants and systematic under-detection of \CM (recall 11\%; see Appendix~\ref{app:llm-variation-eval}). Since \CM accounts for fewer than 2\% of occurrences, this has little impact on aggregate scores.

In our corpora, exact-form matching accounts for about 62--63\% of term occurrences, with lemma matching increasing these numbers to about 82--87\%. Additional matches contributed by \Concordancer\ (static and dynamic variants) are dominated by lexical (\VL) and reduction (\VR) forms (full breakdown in Appendix~\ref{app:term-detection-coverage}).

\paragraph{Alignment.}
We apply Bertalign \citep{liu-etal-2022-bertalign} for sentence-level alignment, and the token-level correspondences are computed with SimAlign \citep{jalili-sabet-etal-2020-simalign}.\footnote{We use the HuggingFace models \href{https://huggingface.co/sentence-transformers/paraphrase-multilingual-mpnet-base-v2}
{\url{paraphrase-multilingual-mpnet-base-v2}} and \href{https://huggingface.co/facebook/xlm-roberta-xl}
{\url{facebook/xlm-roberta-xl}} as the embedding models for Bertalign and SimAlign, respectively.}
For each source term occurrence, the aligned target span is assembled from individual token alignments. When the cosine similarity of the assembled span falls below $0.8$, the bilingual sentence pair, the source term, and the SimAlign candidate are submitted to ChatGPT (\texttt{gpt-4.1-mini}; prompt in Appendix~\ref{app:alignment-prompt}); the LLM returns the translated form of the source term as it appears in the target sentence (an empty return signals an omission), which replaces the SimAlign span. The $0.8$ cutoff is a conservative operating point that reserves the LLM fallback for spans where the embedding aligner is least confident, keeping higher-similarity spans unchanged.
To validate this step, the same annotator evaluated 100~term alignments sampled across the nine system outputs; 96\% (96/100) were correct, with the four errors limited to span-boundary and lexical-choice mistakes on single occurrences (see Appendix~\ref{app:alignment-eval}).

\paragraph{Target-side labeling.}
Target-side analysis combines two parallel processes that are reconciled afterwards. (1)~\Concordancer\ scans the full target document and labels each detected occurrence using the same two-level protocol as the source side (\Concordancer\ + ChatGPT), producing a variation label relative to the expected target preferred term. (2)~The alignment step assembles, for each source term occurrence, a target span. We reconcile the two by matching each canonicalized aligned span against \Concordancer's target term list: all aligned spans found by \Concordancer\ inherit its variation label, while aligned spans missed by \Concordancer\ are directly labeled by ChatGPT, given the source term and the aligned span. Each realization is additionally assigned a glossary conformity label: \textit{glossary-conforming}, \textit{variant translation}, or \textit{out-of-glossary} (paraphrase, omission, or error).

\section{Evaluation Framework}
\label{sec:evaluation-framework}

We analyse source-target associations of terms along three measurements: \textit{term-level accuracy} (Dimension~1, building on glossary-based exact match \citep{alam2021evaluationmachinetranslationterminology}), \textit{translation consistency} (Dimension~2, building on the consistency tradition \citep{itagaki-etal-2007-automatic,lyu-etal-2021-encouraging,semenov-bojar-2022-automated}), and \textit{cross-term variation} (Dimension~3, introduced here). These quantities serve a dual purpose. Applied to the human translation references, they describe terminological behaviour in a corpus; applied to MT outputs against the source and the reference,
they evaluate translation quality. The descriptive use establishes that the human reference is not arbitrary; the evaluative use asks whether MT systems resembles human decisions, and whether such similarities are even desirable.

\paragraph{Dimension 1: Term-level accuracy.}
We compute Exact Match (EM) accuracy: the proportion of occurrences whose translation matches the glossary entry. EM can be applied to any occurrence whose canonical form maps to a glossary entry (preferred term or alternative term).

\paragraph{Dimension 2: Translation consistency.}
For each canonical surface form (that is, each preferred term or variant treated independently) with at least two occurrences, we collect all its target counterparts in a document and measure the consistency of the resulting translation set. We apply three metrics to compute a translation consistency ratio (\ITC); none of them account for whether the target-side variation is justified by source-side differences.

\begin{itemize}
    \item \textbf{\ITCnoxspace$_{\text{EM}_{\text{macro}}}$ (\ITC-EM)}: for each term with $k \geq 2$ occurrences, computes the proportion of occurrences that match the glossary, then averages across terms (term-level).
    \item \textbf{\ITC temporal (\ITC-T)}: for each pair of successive occurrences of the same source canonical form (the 1st and 2nd, then the 2nd and 3rd, etc.), \ITC-T checks whether they receive identical translations. The score is the proportion of consecutive pairs that agree.
    \item \textbf{\ITC pseudo-ref (\ITC-PR)}: consistency measured against a pseudo-reference form (the first occurrence). This score does not require a glossary and applies even when a terminological resource is lacking.
\end{itemize}
Of these, \ITC-PR implements the pseudo-reference consistency of \citet{semenov-bojar-2022-automated}, \ITC-EM applies glossary-based EM \citep{alam2021evaluationmachinetranslationterminology} at the term level, and \ITC-T is a temporal variant we adopt to capture whether translation shifts are gradual or repeated; we retain all three, rather than select one, because their divergence is itself a finding: the choice of consistency definition is not neutral and can change system rankings (Section~\ref{ssec:metric-penalties}).
\ITC-EM combines Dimensions~1 and~2: occurrences must agree with each other and with the glossary. This makes it more restrictive than \ITC-T and \ITC-PR: a term consistently translated with a correct but non-glossary variant translation would score 0 despite being perfectly consistent. \ITC-T and \ITC-PR are purely consistency-oriented: they measure whether a term is translated uniformly, regardless of which translation is used. Table~\ref{tab:tlc-example} (Appendix~\ref{app:tlc-example}) illustrates how these metrics can diverge on the same data.

\paragraph{Dimension 3: Cross-term variation (\CTV).}
While the previous dimensions assess each preferred term or variant independently (each treated as a separate canonical surface form), this dimension evaluates whether the variation relationships between a preferred term and its variants are transferred to the target. Specifically, we measure \textit{Variation preservation}: when a source preferred term is correctly translated, each \VG/\VE/\VR/\VL variant should be translated as a corresponding form in the target. We retain these four types because each has a predictable expected target form: a source acronym is typically rendered as a target acronym (\VG), a truncated source form as a target truncation (\VR), an expanded source form as a target expansion (\VE), and a near-synonymous substitution as a target lexical variant or the target preferred term itself (\VL).\footnote{\VMS and \CM are excluded because their source-to-target mapping is less predictable: morphosyntactic changes depend on target-language grammar, and compound variants involve multiple simultaneous transformations.}

\begin{table}[H]
\centering
\footnotesize
\setlength{\tabcolsep}{4pt}
\resizebox{\columnwidth}{!}{%
\begin{tabular}{p{1.6cm} p{3.6cm} p{2.2cm}}
\toprule
\textbf{Metric} & \textbf{Definition} & \textbf{Exclusions} \\
\midrule
\multicolumn{3}{l}{\textbf{1. Term-Level Accuracy}} \\
\midrule
EM (micro)
& \textit{Occurrence-level conformity}
& OOG terms \\
\midrule
\multicolumn{3}{l}{\textbf{2. Translation Consistency}} \\
\midrule
\TLC-PR
& \textit{Match to first occurrence}
& $k_t < 2$ \\
\TLC-T
&  \textit{Consecutive stability}
& $k_t < 2$ \\
\TLC-EM
& \textit{Avg.\ conformity per term}
& $k_t < 2$ + OOG \\
\midrule
\multicolumn{3}{l}{\textbf{3. Cross-Term Variation}} \\
\midrule
\CTV
& \textit{Preserve variation type}
& \VMS,\CM \\
\bottomrule
\end{tabular}%
}
\caption{Evaluation dimensions. $k_t$: number of occurrences of term $t$. OOG: out-of-glossary.}
\label{tab:metrics_overview}
\end{table}

Table~\ref{tab:metrics_overview} summarizes all metrics, their objectives, and exclusion criteria. Importantly, the framework is diagnostic: it characterizes where and how current evaluation clashes with term variation but does not assess translation quality. A term may be consistently translated yet semantically wrong, or inconsistently translated in a contextually appropriate manner.

\section{Experimental Setup}
\label{sec:setup}

We conduct our analysis on two English--French parallel corpora in the NLP domain: \ACL and \iwslt. Table~\ref{tab:termvar-stats} summarizes their statistics.
\paragraph{\ACL}
The \ACL corpus \citep{peng-etal-2026-parallel} consists of 32 English papers published in NLP conferences (2002--2015) paired with comparable French versions published in TALN\footnote{Traitement Automatique des Langues Naturelles, \url{https://www.atala.org/-Conference-TALN-RECITAL}.} or JEP\footnote{Journées d'Etude de la Parole, \url{https://www.afcp-parole.org/category/animations-scientifiques/les-jep/}.} conferences, resulting in 6,272 parallel sentences. It was constructed via automatic sentence alignment, with Automatic Post-Editing and MT used to handle poorly aligned segments (details in Appendix~\ref{app:acl-corpus}).
\paragraph{\iwslt}
The \iwslt corpus comprises the validation and evaluation sets of the IWSLT 2023 translation shared task \citep{salesky-etal-2023-evaluating}. %\footnote{International Workshop on Spoken Language Translation, \url{http://iwslt.org}.} 
It contains manually revised transcripts and translations of 10 presentations delivered at the ACL 2022 conference,\footnote{\url{https://2022.aclweb.org}.} aligned at the sentence level. We use the English--French subset.

\paragraph{Terminological resource.}
Our terminological reference is a manually curated bilingual glossary covering 1,642 NLP concepts with their preferred English and French realizations, compiled from domain-specific resources and validated by experts. It was prepared and published by INIST, the French Institute for Scientific and Technological Information.\footnote{\url{https://skosmos.loterre.fr/8LP/fr/}}

\paragraph{Models and Inference.}
We evaluate four medium-size multilingual LLMs using vLLM \citep{kwon2023efficient} (model cards in Appendix~\ref{app:model-cards}): Llama3.1-8B-Instruct (Llama) \citep{grattafiori-etal-2024-llama3}, Qwen3-8B (Qwen) \citep{yang-etal-2025-qwen3}, EuroLLM-9B-Instruct (Euro9) \citep{martins-etal-2025-eurollm9b}, and EuroLLM-22B-Instruct (Euro22) \citep{ramos-2026-etal-eurollm22b}. We refer to the human reference as ``Ref''.
We apply greedy decoding to all systems except Qwen, for which we use the recommended ``non-thinking'' mode (temperature=$0.7$, top-$p$=$0.8$, top-$k$=$20$).

\paragraph{Translation settings.}
We evaluate MT outputs under two zero-shot conditions: (1)~\MTbaseline, where models translate (English-to-French) without terminological guidance; and (2)~glossary-guided (\MTmoslem), where glossary-preferred terms and their expected translations are injected into the prompt. Comparing the two conditions serves as a diagnostic for whether constrained systems genuinely exploit the glossary information (prompts in Appendix~\ref{app:translation-prompt}).

\section{Analysis}
\label{sec:analysis}

\subsection{Variation Patterns: Reference and MT}
\label{ssec:reference-variation}

\begin{table}[H]
    \centering
    \small
    \resizebox{\columnwidth}{!}{%
    \begin{tabular}{lrr}
    \toprule
    \textbf{Statistic} & \textbf{\ACL} & \textbf{\iwslt} \\
    \midrule
    \multicolumn{3}{l}{\textit{Corpus overview}} \\[2pt]
    \#docs
        & 32
        & 10 \\
    Avg.\ \#sents / doc
        & 196
        & 88 \\
    Avg.\ concepts / doc
        & 50.44
        & 31.00 \\
    \midrule
    \multicolumn{3}{l}{\textit{Occurrence \& variation richness}} \\[2pt]
    Avg.\ occ.\ / concept
        & 5.55
        & 3.92 \\
    Avg.\ distinct terms / concept
        & 1.37
        & 1.14 \\
    \midrule
    \multicolumn{3}{l}{\textit{Concept distribution}} \\[2pt]
    Single-term concepts (\% of concepts)
        & 77.74
        & 88.27 \\
    Single-term concepts (\% of occ.)
        & 64.44
        & 76.94 \\
    Hapax concepts (\% of concepts)
        & 37.56
        & 42.68 \\
    Hapax concepts (\% of occ.)
        & 7.58
        & 11.55 \\
    \midrule
    \multicolumn{3}{l}{\textit{Variation category breakdown (\% of occ.)}} \\[2pt]
    \NV
        & 72.28
        & 74.51 \\
    \VL
        & 7.97
        & 5.66 \\
    \VR
        & 9.21
        & 8.93 \\
    \VE
        & 1.10
        & 0.39 \\
    \VMS
        & 0.59
        & 0.29 \\
    \VG
        & 7.54
        & 8.73 \\
    \CM
        & 1.32
        & 1.49 \\
    \bottomrule
    \end{tabular}%
    }
    \caption{Terminological variation statistics of the source texts,
    averaged at the document level.
    }
    \label{tab:termvar-stats}
\end{table}

Table~\ref{tab:termvar-stats} characterizes source-side terminological variation in both corpora. Each concept appears under 1.37 distinct surface forms in \ACL and 1.14 in \iwslt on average. 77.7\% and 88.3\% of concepts respectively are realized by a single form (full distribution in Appendix~\ref{app:distinct-terms-distribution}). Variation is concentrated among frequent concepts, and preferred terms make up 72--75\% of occurrences. Among variation types, \VR and \VG account for 7--9\% each, \VL for 6--8\%, and \VMS, \VE, and \CM together represent less than 3\%. Since 38--43\% of concepts are hapaxes, any score requiring multiple occurrences is blind to much of the terminology.

The analysis of reference targets in this same setting (Table~\ref{tab:variation-comparison-null_replace-occ_gt1-reorg}, row \emph{Ref}) confirms that variation is not random noise: most concepts retain their number of distinct surface forms across languages. 
However, in \ACL, 34.4\% of concepts exhibit more target-side than source-side realizations, higher than the MT systems. In \iwslt\ (spoken transcripts), the rate drops to 15.8\%. This register asymmetry is consistent with the well-documented tendency of French academic writing to favor lexical variation (\emph{variation élégante}) \citep{vinay-darbelnet-1995}.

\begin{table}[H]
\centering\small
\setlength{\tabcolsep}{3pt}
\begin{tabular}{ll ccccc}
\toprule
\textbf{System} & \textbf{Var.} & Ref & E22 & E9 & Qwen & Llama \\
\midrule
\multicolumn{7}{l}{\textbf{\ACL}} \\
\midrule
\multirow{2}{*}{\MTbaseline}
& $=$ & 63.7 & 69.1 & 68.5 & 66.5 & 66.6 \\
& $<$ & 34.4 & 29.4 & 29.8 & 32.2 & 31.9 \\
\multirow{2}{*}{\MTmoslem}
& $=$ & 63.8 & 73.8 & 72.4 & 73.2 & 73.6 \\
& $<$ & 34.3 & 24.1 & 25.3 & 24.6 & 24.8 \\
\midrule
\multicolumn{7}{l}{\textbf{\iwslt}} \\
\midrule
\multirow{2}{*}{\MTbaseline}
& $=$ & 82.5 & 67.8 & 66.7 & 69.5 & 64.4 \\
& $<$ & 15.8 & 31.6 & 32.8 & 29.9 & 35.0 \\
\multirow{2}{*}{\MTmoslem}
& $=$ & 82.5 & 77.4 & 74.0 & 79.7 & 77.4 \\
& $<$ & 15.8 & 21.5 & 24.9 & 19.2 & 21.5 \\
\bottomrule
\end{tabular}
\caption{Distribution (\%) of concepts by source/target variation ($=$: EN and FR have the same number of distinct terms; $<$: FR has more). Row EN$>$FR omitted (all values below 2.3\%). Avg.\ concepts/doc: 31.48 (\ACL), 17.70 (\iwslt).}
\label{tab:variation-comparison-null_replace-occ_gt1-reorg}
\end{table}

Human translators also routinely replace preferred terms with lexical variants: in the reference, source preferred terms are rendered by a target-side lexical variant (\VL) in 24.98\% of cases in \iwslt and 15.68\% in \ACL (full transfer matrices in Appendix~\ref{app:transfer-matrices}). A concrete case: within a single sentence of the \ACL corpus, the English term \textit{reported speech} receives two distinct French translations in the reference (\textit{parole rapportée} and \textit{discours rapporté}, both valid translations of the same phenomenon), a stylistic alternation that a consistency metric may penalize as inconsistency (full example with MT outputs in Table~\ref{tab:reported-speech-variation}, Appendix~\ref{app:reported-speech-variation-extended}).

\textbf{MT systems reproduce variation unevenly.}\label{ssec:variation-transfer}
Whether MT outputs reproduce the source's variation depends on register and variation type. For academic prose (\ACL), MT amplification (29--33\%) is in the same ballpark as the reference (34.4\%); for transcripts (\iwslt), MT systems amplify more (30--35\%) than the reference (15.8\%). %No MT system matches the reference across both registers.

The differences deepen when we break down variation transfer by type, assessing how each source variation type (preferred term, acronym, reduction, etc.) is rendered on the target side (full transfer matrices in Appendix~\ref{app:transfer-matrices}). \VR transfers most reliably, with 83--86\% preservation on both corpora. \VG, in contrast, shows the widest gap between corpora: 50--60\% preservation on \ACL drops to 14--29\% on \iwslt, where source acronyms are far more often spelled out as a French phrase than kept as an abbreviation. This likely reflects the translation guidelines of \iwslt rather than the spoken source: even the human reference spells acronyms out more than the MT systems, which a source-side effect could not explain. \NV preservation is likewise higher on \ACL (81--85\%) than on \iwslt (71--75\%). Evaluating all variation types uniformly therefore conflates phenomena with fundamentally different cross-lingual behaviors. For instance, in a single \ACL document the acronym \textit{RL} (\VG) for \textit{reinforcement learning} is preserved in one context but expanded into different forms in others (full example in Table~\ref{tab:rl-ctv-example}, Appendix~\ref{app:rl-ctv-example}).

\subsection{Current Metrics Penalize Valid Variation}
\label{ssec:metric-penalties}

\begin{table}[H]
\centering\small
\setlength{\tabcolsep}{3pt}
\begin{tabular}{@{}l rr rr@{}}
\toprule
& \multicolumn{2}{c}{\textbf{\ACL}} & \multicolumn{2}{c}{\textbf{\iwslt}} \\
\cmidrule(lr){2-3} \cmidrule(lr){4-5}
Method & occ & dist & occ & dist \\
\midrule
EM           & 85.0 & 71.7 & 90.4 & 83.3 \\
\ITC         & 86.9 & 50.5 & 85.6 & 53.2 \\
\CTV         &  9.4 & 12.5 &  2.7 &  4.2 \\
\midrule
EM+\ITC      & 95.1 & 81.1 & 96.9 & 90.0 \\
EM+\CTV      & 91.9 & 82.6 & 92.8 & 87.2 \\
\ITC+\CTV    & 88.7 & 57.7 & 86.2 & 55.0 \\
\midrule
EM+\ITC+\CTV & 96.8 & 87.9 & 97.5 & 91.9 \\
\bottomrule
\end{tabular}
\caption{Average per-document coverage (\%) by method. \textit{occ}: each occurrence; \textit{dist}: each distinct term counted once. \ITC uses non-EM variants.}
\label{tab:cross_method_coverage}
\end{table}

Before reporting scores for the three measurements, we discuss their \emph{coverage}, i.e.~the fraction of source-side occurrences each one evaluates. The three evaluation dimensions have complementary coverage profiles (Table~\ref{tab:coverage_by_cat_doc_avg}). EM covers 100\% of \NV terms by design but this drops sharply for variants.

EM's coverage drop on variants is structural: \VMS and \CM are essentially invisible to glossary-based evaluation, and \VG coverage varies across corpora (97\% in \iwslt, where most acronyms are listed in the glossary, versus 59\% in \ACL, where many domain-specific acronyms are absent from it). \ITC covers 86--87\% of occurrences on both corpora but applies only to recurrent canonical forms, assessing each surface form independently and missing cross-variant coherence. \CTV has the narrowest raw coverage but provides a qualitatively distinct signal: it is the only dimension that evaluates hapax variants (38--43\% of concepts) through their structural relationship to the preferred term. Combined, the three dimensions reach 96.8\% of occurrences in \ACL and 97.5\% in \iwslt, leaving a small structurally irreducible residual (Table~\ref{tab:cross_method_coverage}; details in Appendix~\ref{app:unevaluated}).

Table~\ref{tab:strategy-baseline} reports the three-dimensional scores for MT systems (\MTbaseline and \MTmoslem); we focus on the \MTbaseline\ here, leaving \MTmoslem\ for Section~\ref{ssec:glossary-tension}. Aggregate EM places all systems within a narrow range per corpus, hiding a structural effect: the per-variation breakdown (Appendix~\ref{app:scores-breakdown}) shows that aggregate EM is driven by preferred-term performance (72--85\% on \NV) while variant scores are substantially lower. The aggregate therefore says little about how systems handle variants.

\begin{table}[H]
\centering\scriptsize
\setlength{\tabcolsep}{2pt}
\renewcommand{\arraystretch}{0.9}
\resizebox{\columnwidth}{!}{%
\begin{tabular}{l rcccc rcccc}
\toprule
& \multicolumn{5}{c}{\textbf{\ACL}} & \multicolumn{5}{c}{\textbf{\iwslt}} \\
\cmidrule(lr){2-6} \cmidrule(lr){7-11}
Cat. & $\bar{N}$ & EM & \ITC & \ITC-EM & \CTV & $\bar{N}$ & EM & \ITC & \ITC-EM & \CTV \\
\midrule
\NV  & 198 & 100 & 90 & 90 & ---  & 88 & 100 & 87 & 87 & --- \\
\VR  & 25  & 43  & 89 & 42 & 32   & 10 & 50  & 85 & 47 & 17  \\
\VMS & 3   & 8   & 2  & 2  & ---  & 1  & 0   & 0  & 0  & --- \\
\VE  & 4   & 26  & 32 & 13 & 40   & 4  & 42  & 42 & 42 & 58  \\
\VG  & 21  & 59  & 69 & 53 & 32   & 11 & 97  & 82 & 82 & 7   \\
\VL  & 22  & 35  & 63 & 24 & 32   & 8  & 41  & 46 & 22 & 8   \\
\CM  & 5   & 5   & 26 & 5  & ---  & 4  & 0   & 28 & 0  & --- \\
\bottomrule
\end{tabular}%
}
\caption{\% of occurrences evaluated per category, averaged over documents. $\bar{N}$: avg.\ occurrences. Values rounded to the nearest integer.}
\label{tab:coverage_by_cat_doc_avg}
\end{table}

Consistency scores expose the cost of penalizing acceptable variation. The reference dominates for \iwslt under all \ITC variants; for \ACL, differences narrow. The strictest variant (\ITC-EM) results in lower scores across the board, reflecting the compounding difficulty of  requiring both accuracy and consistency.

Consistency is hardest to achieve precisely for terms that inherently exhibit variation: \ITC scores for \VL and \VG are substantially lower than for \NV (Appendix~\ref{app:scores-breakdown}). For instance, the term \textit{pitch} (glossary: \textit{tonie}) receives two contextually appropriate French equivalents (\textit{hauteur}, \textit{m\'elodique}) across 14~occurrences in a single \ACL document, and a consistency metric severely penalizes this justified variation (full example in Table~\ref{tab:pitch-consistency}, Appendix~\ref{app:pitch-consistency}).

\begin{table}[H]
\centering
\scriptsize
\setlength{\tabcolsep}{3pt}
\renewcommand{\arraystretch}{1}
\resizebox{\columnwidth}{!}{%
\begin{tabular}{lccccc|ccccc}
\toprule
& \multicolumn{5}{c|}{\textbf{\ACL}} & \multicolumn{5}{c}{\textbf{\iwslt}} \\
\cmidrule(lr){2-6}\cmidrule(l){7-11}
Metric & Ref & E22 & E9 & Q. & Ll. & Ref & E22 & E9 & Q. & Ll. \\
\midrule
\multicolumn{11}{@{}l}{\textit{\MTbaseline}} \\
\midrule
EM Acc. & \textbf{81.5} & 79.3 & 79.9 & 77.8 & 78.4 & 72.3 & 71.3 & 72.0 & \textbf{74.1} & 69.9 \\
\TLC-EM & \textbf{83.0} & 80.8 & 81.3 & 79.4 & 79.8 & 74.2 & 73.2 & 73.5 & \textbf{76.5} & 72.0 \\
\TLC-PR & 86.7 & 88.2 & 89.2 & 86.6 & \textbf{89.3} & \textbf{95.1} & 84.4 & 87.9 & 88.1 & 86.1 \\
\TLC-T  & 85.7 & 87.2 & \textbf{87.4} & 85.8 & 87.7 & \textbf{94.0} & 83.3 & 84.0 & 84.2 & 82.7 \\
\CTV    & 66.1 & 72.8 & 72.7 & 71.7 & \textbf{74.4} & \textbf{86.7} & 85.7 & 85.7 & 68.2 & 75.8 \\
\midrule
\multicolumn{11}{@{}l}{\textit{\MTmoslem}} \\
\midrule
EM Acc. & 81.5 & 87.9 & 85.4 & 88.3 & \textbf{88.8} & 72.3 & 84.5 & 78.4 & \textbf{87.4} & 86.6 \\
\TLC-EM & 83.0 & 89.0 & 86.4 & 89.2 & \textbf{89.7} & 74.3 & 86.0 & 79.6 & \textbf{89.0} & 87.6 \\
\TLC-PR & 86.8 & \textbf{91.6} & 90.5 & 91.3 & 90.7 & \textbf{95.1} & 89.9 & 90.3 & 93.0 & 91.6 \\
\TLC-T  & 85.7 & \textbf{90.7} & 89.7 & 90.6 & 89.8 & \textbf{94.0} & 87.5 & 88.3 & 90.1 & 89.0 \\
\CTV    & 66.1 & 67.0 & 68.5 & \textbf{69.2} & 68.2 & \textbf{86.7} & 74.2 & 81.0 & 80.6 & 81.1 \\
\bottomrule
\end{tabular}%
}
\caption{Scores for \MTbaseline\ and \MTmoslem. E22/E9=Euro22/Euro9, Q.=Qwen, Ll.=Llama, Bold=highest across systems per metric and corpus.}
\label{tab:strategy-baseline}
\end{table}

The reference's \CTV score makes the same point from a different angle. While the reference achieves the highest \CTV for \iwslt, it scores the lowest for \ACL. This paradox is explained by the higher target-side variation the reference introduces for \ACL, which creates more opportunities for structural coherence failure.
Consistency metrics are not stable across definitions.
For \ACL, each \ITC variant ranks a different system as top-1 (Table~\ref{tab:strategy-baseline}): no single system dominates consistency across all three operationalizations, so the choice of metric is not a neutral implementation detail.

\subsection{Glossary Constraints vs.\ Acceptable Variations}
\label{ssec:glossary-tension}

The \MTmoslem\ block of Table~\ref{tab:strategy-baseline} reports MT scores where the preferred terms from the glossary and their expected translations are injected into the translation prompt. Comparing with the \MTbaseline, EM and \ITC improve substantially for all MT systems: when prompted with the glossary entries, systems readily produce them, showing that they were capable of generating these forms but, without the prompt, often selected alternative wordings. The constraint's effect on the transfer matrices (Appendix~\ref{app:transfer-matrices}) is selective: \NV$\to$\NV increases by 8--16 points while \NV$\to$\VL drops correspondingly, and other variation types are largely unaffected.

\CTV, however, drops for all MT systems under the glossary constraint. Forcing glossary-preferred forms improves both accuracy and consistency at the price of suppressing acceptable target-side variation: the very behavior that Section~\ref{ssec:reference-variation} showed to be a hallmark of human translation. This is the central tension exposed by our framework: no single dimension captures the full picture, and a glossary-constrained system that appears to ``improve'' on EM and \ITC may simultaneously regress on \CTV. The methodological takeaway is that variation-aware evaluation should condition consistency penalties on whether target-side variation mirrors source-side variation, rather than penalizing all deviation uniformly. Mirroring source variation is a sufficient but not a necessary condition for legitimate target variation: it validates the structural case that \CTV\ measures, but not the stylistic alternation of a term that is itself invariant in the source (e.g., \textit{pitch}, \textit{reported speech}), which has no source counterpart to mirror and can only be adjudicated semantically. \CTV\ is therefore diagnostic rather than prescriptive: it flags where surface consistency and structurally legitimate variation diverge, without licensing every target-side deviation.

\section{Conclusion}
\label{sec:conclusion}
Terminological variation is a standard property of scientific translation. However, existing automatic metrics systematically penalize it. Analyzing two English--French NLP corpora translated by four medium-size LLMs, we show that human translators introduce more target-side variation than MT systems on academic prose, and that this variation is often semantically motivated. MT systems reproduce it unevenly across registers, less the reference for academic prose and more for transcripts.

Current automatic metrics handle this variation poorly: glossary-based EM is dominated by preferred-term performance, translation consistency penalizes the acceptable diversification that human translators produce, and different operationalizations of consistency result in divergent system rankings. Cross-term variation (\CTV), the structural coherence dimension we introduce, recovers signal where the other dimensions are blind. No single metric captures terminology quality, motivating variation-aware evaluation that conditions consistency penalties on whether target-side variation mirrors structurally legitimate source-side variation.

The most striking result is that enforcing MT systems to use glossary-preferred forms improves accuracy and consistency but degrades \CTV, by suppressing the acceptable target-side variation that the unconstrained baseline preserved. This reveals the need for better MT engines that produce high-quality translations with terminological accuracy, consistency and acceptable variation.

\section{Limitations}

We carry out our analysis on one language pair (English--French), one domain (NLP scientific text), two  corpora, and medium-size LLMs (8B--22B parameters) evaluated under zero-shot conditions. Variation transfer patterns are likely to differ for typologically distant language pairs (e.g., English--Chinese), where acronym and reduction behavior is fundamentally different, and for other specialized domains with distinct terminological profiles. Larger frontier models or translation-specialized systems such as Tower \citep{alves-etal-2024-tower} may also exhibit different variation behavior; in particular, more capable models might integrate glossary guidance with less suppression of valid variation, an open question that our zero-shot medium-size setting cannot settle. Confirming the generalizability of our findings requires broader experimental coverage.

Our cross-term variation diagnostic (\CTV; Dimension~3, Section~\ref{sec:evaluation-framework}) measures structural coherence under an idealized transfer assumption: it expects each source variation type to be preserved in translation, but legitimate cross-lingual adaptation may well produce a different type (e.g., expanding an acronym for clarity in the target language).

We do not test how neural or reference-free metrics (e.g., COMET, MetricX, or LLM-as-judge) treat terminological variation. Since they score semantic adequacy rather than surface similarity, they may be more robust to legitimate alternation. However, they commonly operate at the segment level and are not well adapted for the document level. Whether they can also diagnose the terminological inconsistency that our surface-level dimensions target therefore remains an open question.

Several pipeline choices condition our results. The variation classifier uses a closed commercial LLM (\texttt{gpt-4.1-mini}), so exact reproducibility depends on API-version stability; manual validation on a sample of 50 pairs reports 84\% type accuracy, with reduced recall on compound variants (11\%). As \CM accounts for fewer than 2\% of all occurrences, this affects only finer-grained per-type conclusions. Using an LLM to produce variation labels that are then used to evaluate other LLMs' translations introduces a potential circular dependency, although the tasks involved (classification vs.\ translation) are distinct. Alignment errors from Bertalign and SimAlign propagate to all three evaluation dimensions; a manual audit of 100 sampled alignments estimates their rate at 4\% (Appendix~\ref{app:alignment-eval}).
Finally, \Concordancer\ depends on language-specific resources (spaCy, MorphyNet) and rule-based heuristics for dynamic variant detection, which may miss unconventional syntactic constructions. We quantify what \Concordancer\ finds (Appendix~\ref{app:term-detection-coverage}) but not what it misses, as no gold-standard annotation of all term occurrences exists to measure detection recall. The framework also requires a bilingual glossary, which may not be available for all domains or language pairs.

\section*{Acknowledgements}
This work was supported by the French national research agency (ANR) as part of the MaTOS project (grant number: ANR-22-CE23-0033).\footnote{\url{http://anr-matos.fr/}} Rachel Bawden was also partly funded by her chair position in the PRAIRIE institute funded by ANR as part of the ``Investissements d'avenir'' programme under reference ANR19-P3IA-0001. The authors are grateful to the anonymous reviewers for their insightful comments and suggestions.

\begin{small}
\bibliographystyle{apalike}
\bibliography{custom}
\end{small}

\appendix
\section{Appendix}
\label{sec:appendix}

\subsection{Term Detection Coverage}
\label{app:term-detection-coverage}

\begin{table}[H]
\centering\small
\setlength{\tabcolsep}{6pt}
\renewcommand{\arraystretch}{0.95}
\begin{tabular}{lrr}
\toprule
                       & \ACL   & \iwslt \\
\midrule
Exact-form match (\%)  & 61.9\% & 62.7\% \\
Lemma match (\%)       & 82.2\% & 86.6\% \\
Avg.~terms / doc       & 273.8  & 120.3  \\
\bottomrule
\end{tabular}
\caption{\Concordancer\ recognition coverage, averaged per document. Exact-form and lemma matching are baselines that a naive identification method could use; \Concordancer\ recovers the remaining terms by detecting further variants.}
\label{tab:concordancer-recognition}
\end{table}

\begin{table}[H]
\centering\small
%\resizebox{\linewidth}{!}{
\setlength{\tabcolsep}{4pt}
\renewcommand{\arraystretch}{0.95}
\begin{tabular}{lrrrr}
\toprule
 & \multicolumn{2}{c}{\ACL} & \multicolumn{2}{c}{\iwslt} \\
\cmidrule(lr){2-3} \cmidrule(lr){4-5}
 & \multicolumn{1}{c}{ING} & \multicolumn{1}{c}{OOG} & \multicolumn{1}{c}{ING} & \multicolumn{1}{c}{OOG} \\
\midrule
Avg.~terms / doc & 232.8  & 40.9   & 109.1  & 11.2   \\
\midrule
\NV  & 84.3\% & --     & 82.5\% & --     \\
\VG  & 7.0\%  & 11.3\% & 9.2\%  & 3.9\%  \\
\VL  & 3.5\%  & 35.7\% & 3.2\%  & 31.3\% \\
\VE  & 0.5\%  & 5.0\%  & 0.3\%  & 1.7\%  \\
\VR  & 4.6\%  & 36.3\% & 4.8\%  & 45.5\% \\
\VMS & 0.0\%  & 4.1\%  & 0.0\%  & 2.7\%  \\
\CM  & 0.1\%  & 7.6\%  & 0.0\%  & 14.9\% \\
\bottomrule
\end{tabular}%}
\caption{Distribution of variation types among \Concordancer-detected terms, averaged per document and split by whether the term's lemma is listed in the glossary (\textit{ING}, in-glossary) or not (\textit{OOG}, out-of-glossary). Percentages are within each group.}
\label{tab:concordancer-coverage}
\end{table}

Table~\ref{tab:concordancer-recognition} reports the recognition coverage of \Concordancer\ on both corpora. Exact-form and lemma matching (the two naive identification baselines) would respectively retrieve 61.9\% and 82.2\% of terms for \ACL, and 62.7\% and 86.6\% for \iwslt. \Concordancer, which detects additional variant types, identifies the remaining terms.

Table~\ref{tab:concordancer-coverage} breaks down \Concordancer-detected terms into \textit{in-glossary} terms (\textit{ING}), whose lemma appears in the glossary, and \textit{out-of-glossary} terms (\textit{OOG}), discovered by \Concordancer\ beyond its scope. For \textit{ING} terms, more than 80\% are preferred terms (\NV). For \textit{OOG} terms, the distribution shifts: \VL and \VR together account for more than 70\%, meaning \Concordancer\ primarily discovers out-of-glossary terms through lexical and reduction variants. Since no glossary reference exists for \textit{OOG} terms, their translation cannot be assessed by term-level accuracy (Dimension~1), motivating the complementary evaluation dimensions.

\subsection{Distribution of Distinct Terms per Concept}
\label{app:distinct-terms-distribution}

\begin{figure}[H]
  \centering
  \includegraphics[width=0.8\linewidth]{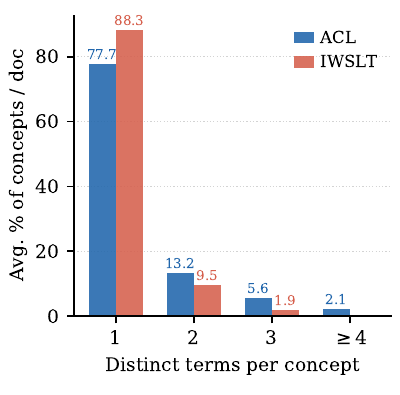}
  \caption{Distribution of the number of distinct terms per concept, averaged by document, showing the proportion of concepts with exactly $k$ distinct terms.}
\label{fig:distinct-terms-dist}
\end{figure}

Figure~\ref{fig:distinct-terms-dist} reports the distribution of the number of distinct canonical surface forms (preferred term and variants) per concept, averaged at the document level. The distribution is strongly skewed: most concepts are realized by a single form in both corpora, with a long tail of concepts showing two or more variants.

\subsection{Variation Transfer Matrices}
\label{app:transfer-matrices}

Tables~\ref{tab:var-transfer-baseline-ACL-simalign-output} and~\ref{tab:var-transfer-baseline-IWSLT-simalign-output} present the full source-to-target variation transfer matrices under the \MTbaseline condition. Each cell reports the percentage of source occurrences of a given variation type (rows) that are translated as a given target variation type (columns). Diagonal values indicate type preservation.
Key observations beyond those discussed in Section~\ref{ssec:variation-transfer}:
\begin{itemize}
    \item \textbf{\NV$\to$\NV}: 81--85\% on \ACL but only 71--75\% on \iwslt. The off-diagonal \NV$\to$\VL is the main source of preferred-term variation: 10--12\% on \ACL and 18--25\% on \iwslt, with the reference showing the highest rate on both corpora.
    \item \textbf{\VG$\to$\VG}: 50--60\% on \ACL but only 14--29\% on \iwslt, where 63--79\% of source \VG are realized as \VR instead. This likely reflects the translation guidelines of \iwslt, where source acronyms are systematically spelled out rather than kept as an abbreviation (the human reference does so even more than the MT systems), rather than the spoken origin of the source.
    \item \textbf{\VR$\to$\VR}: 83--86\% on both corpora, the most stable transfer across all types and systems.
    \item \textbf{\VL$\to$\VL}: 58--63\% on \ACL and 52--77\% on \iwslt, with the reference achieving notably higher preservation than MT systems on \iwslt.
\end{itemize}

Tables~\ref{tab:var-transfer-moslem-ACL-simalign-output} and~\ref{tab:var-transfer-moslem-IWSLT-simalign-output} present the same matrices under the \MTmoslem condition. The glossary constraint selectively affects preferred-term transfer: \NV$\to$\NV increases from 81--84\% to 90--93\% on \ACL and from 71--75\% to 79--91\% on \iwslt, while \NV$\to$\VL drops from 10--12\% to 3--5\% on \ACL and from 18--23\% to 5--16\% on \iwslt. Other variation types are largely unaffected: \VR$\to$\VR remains at 84--86\% and \VG$\to$\VG at 49--54\% on \ACL, indicating that the glossary constraint acts primarily on preferred terms and does not propagate to variant categories.

\begin{table*}
  \centering
  \footnotesize
  \setlength{\tabcolsep}{4pt}
  \renewcommand{\arraystretch}{0.95}
  \begin{tabular}{ll|rrrrrrrr}
    \toprule
    System & Src & \NV & \VL & \VR & \VG & \VE & \VMS & \CM & $\emptyset$ \\
    \midrule
    \multirow{7}{*}{\textbf{Ref.}}
      & \NV  & \textbf{85.2} & 9.9  & 1.5  & 0.9  & 0.1  & 2.4  & 0.0 & 0.1 \\
      & \VL  & 7.1  & \textbf{63.0} & 18.7 & 4.1  & 1.8  & 4.1  & 0.4 & 0.9 \\
      & \VR  & 0.1  & 7.4  & \textbf{86.2} & 2.4  & 0.6  & 1.4  & 0.2 & 1.7 \\
      & \VG  & 13.2 & 13.6 & 9.6  & \textbf{49.9} & 5.4  & 3.9  & 3.1 & 1.1 \\
      & \VE  & 9.0  & 14.4 & 9.6  & 13.9 & \textbf{26.1} & 16.4 & 7.9 & 2.8 \\
      & \VMS & 12.3 & 24.1 & 17.6 & 0    & 10.2 & \textbf{35.8} & 0   & 0 \\
      & \CM  & 7.2  & \textbf{34.2} & 14.0 & 13.4 & 5.7  & 22.7 & 1.9 & 0.9 \\
    \midrule
    \multirow{7}{*}{\textbf{Euro22}}
      & \NV  & \textbf{83.0} & 10.9 & 1.6  & 0.7  & 0.1  & 3.6  & 0   & 0.0 \\
      & \VL  & 6.8  & \textbf{59.5} & 22.9 & 4.8  & 0.9  & 3.9  & 0   & 1.1 \\
      & \VR  & 0    & 9.7  & \textbf{83.4} & 2.2  & 0.2  & 2.7  & 0.1 & 1.6 \\
      & \VG  & 9.5  & 9.9  & 8.4  & \textbf{53.9} & 10.4 & 2.4  & 4.8 & 0.7 \\
      & \VE  & 4.6  & 19.3 & 1.5  & 24.8 & \textbf{31.4} & 7.2  & 8.4 & 2.8 \\
      & \VMS & 13.3 & 14.0 & 17.6 & 0    & 8.9  & \textbf{44.5} & 1.1 & 0.6 \\
      & \CM  & 4.4  & 26.4 & 14.1 & 11.7 & 8.5  & \textbf{31.0} & 2.9 & 0.9 \\
    \midrule
    \multirow{7}{*}{\textbf{Euro9}}
      & \NV  & \textbf{83.5} & 10.4 & 1.6  & 0.9  & 0.1  & 3.4  & 0   & 0.1 \\
      & \VL  & 6.2  & \textbf{61.8} & 21.5 & 4.8  & 1.2  & 3.2  & 0   & 1.3 \\
      & \VR  & 0.2  & 10.2 & \textbf{82.8} & 2.5  & 0.3  & 2.4  & 0   & 1.6 \\
      & \VG  & 6.6  & 7.9  & 7.6  & \textbf{58.9} & 7.4  & 4.3  & 6.3 & 0.9 \\
      & \VE  & 8.5  & 15.5 & 9.9  & 23.4 & \textbf{24.6} & 13.9 & 1.4 & 2.8 \\
      & \VMS & 13.8 & 19.9 & 15.3 & 0    & 7.2  & \textbf{42.6} & 0.6 & 0.6 \\
      & \CM  & 4.0  & 25.7 & 16.3 & 8.7  & 7.3  & \textbf{34.7} & 2.5 & 0.9 \\
    \midrule
    \multirow{7}{*}{\textbf{Qwen}}
      & \NV  & \textbf{81.4} & 12.5 & 1.4  & 1.7  & 0.1  & 2.8  & 0.0 & 0.1 \\
      & \VL  & 6.3  & \textbf{58.8} & 24.9 & 4.0  & 1.6  & 3.1  & 0.1 & 1.3 \\
      & \VR  & 0.1  & 10.2 & \textbf{83.0} & 2.3  & 0.4  & 2.2  & 0.1 & 1.8 \\
      & \VG  & 5.9  & 13.7 & 7.5  & \textbf{57.5} & 7.1  & 0.5  & 7.2 & 0.7 \\
      & \VE  & 2.1  & 23.8 & 2.3  & \textbf{24.2} & 18.8 & 17.9 & 8.2 & 2.8 \\
      & \VMS & 14.6 & 20.5 & 16.7 & 0.6  & 5.7  & \textbf{40.9} & 0.6 & 0.6 \\
      & \CM  & 5.9  & 30.8 & 11.6 & 7.3  & 6.8  & \textbf{34.4} & 2.2 & 0.9 \\
    \midrule
    \multirow{7}{*}{\textbf{Llama}}
      & \NV  & \textbf{81.6} & 11.1 & 1.2  & 2.5  & 0.1  & 3.4  & 0   & 0.0 \\
      & \VL  & 6.0  & \textbf{59.6} & 22.2 & 5.0  & 1.7  & 4.0  & 0.2 & 1.3 \\
      & \VR  & 0.1  & 8.7  & \textbf{85.5} & 2.3  & 0.2  & 1.5  & 0   & 1.7 \\
      & \VG  & 7.7  & 7.4  & 10.5 & \textbf{59.4} & 5.0  & 1.9  & 7.3 & 0.9 \\
      & \VE  & 9.2  & 18.9 & 2.7  & 23.4 & \textbf{32.9} & 7.9  & 2.1 & 2.8 \\
      & \VMS & 10.0 & 27.3 & 12.7 & 0    & 11.4 & \textbf{38.1} & 0   & 0.6 \\
      & \CM  & 4.0  & \textbf{33.4} & 15.1 & 8.1  & 7.3  & 29.1 & 2.2 & 0.9 \\
    \bottomrule
  \end{tabular}
  \caption{Variation transfer from source to target (\%, mean per document) per system, \MTbaseline\ condition, \ACL. Rows: source variation category per system; columns: target variation category. Diagonal (preservation) in bold. Source distribution (ref., mean/doc): \NV: 72.3\% (198), \VL: 7.9\% (22), \VR: 9.0\% (25), \VG: 7.7\% (21), \VE: 1.1\% (4), \VMS: 0.6\% (2), \CM: 1.3\% (5).}
  \label{tab:var-transfer-baseline-ACL-simalign-output}
\end{table*}

\begin{table*}
  \centering
  \footnotesize
  \setlength{\tabcolsep}{4pt}
  \renewcommand{\arraystretch}{0.95}
  \begin{tabular}{ll|rrrrrrrr}
    \toprule
    System & Src & \NV & \VL & \VR & \VG & \VE & \VMS & \CM & $\emptyset$ \\
    \midrule
    \multirow{7}{*}{\textbf{Ref}}
      & \NV  & \textbf{72.6} & 21.1 & 1.4  & 0.8  & 0    & 4.0  & 0   & 0.1 \\
      & \VL  & 3.5  & \textbf{77.2} & 8.1  & 0.6  & 0    & 10.6 & 0   & 0 \\
      & \VR  & 0    & 11.5 & \textbf{86.4} & 0    & 0    & 1.0  & 1.1 & 0 \\
      & \VG  & 0    & 0    & \textbf{78.6} & 14.2 & 2.0  & 5.3  & 0   & 0 \\
      & \VE  & 0    & 41.7 & 0    & 0    & \textbf{58.3} & 0    & 0   & 0 \\
      & \VMS & 0    & 0    & \textbf{66.7} & 0    & 0    & 33.3 & 0   & 0 \\
      & \CM  & 0    & 8.3  & 20.8 & 8.3  & 8.3  & \textbf{45.8} & 8.3 & 0 \\
    \midrule
    \multirow{7}{*}{\textbf{Euro22}}
      & \NV  & \textbf{72.2} & 19.8 & 1.0  & 0.6  & 0    & 6.2  & 0   & 0.1 \\
      & \VL  & 11.5 & \textbf{54.7} & 13.7 & 9.5  & 0    & 10.6 & 0   & 0 \\
      & \VR  & 0    & 15.7 & \textbf{81.2} & 0    & 0    & 2.0  & 1.1 & 0 \\
      & \VG  & 0    & 0    & \textbf{66.7} & 26.6 & 1.0  & 5.3  & 0.4 & 0 \\
      & \VE  & 0    & 25.0 & 0    & 16.7 & \textbf{58.3} & 0    & 0   & 0 \\
      & \VMS & 0    & 0    & \textbf{100.0} & 0   & 0    & 0    & 0   & 0 \\
      & \CM  & 0    & 18.1 & \textbf{37.5} & 8.3 & 8.3  & 19.4 & 8.3 & 0 \\
    \midrule
    \multirow{7}{*}{\textbf{Euro9}}
      & \NV  & \textbf{73.1} & 19.2 & 1.2  & 1.0  & 0    & 5.6  & 0   & 0 \\
      & \VL  & 14.9 & \textbf{58.6} & 8.1  & 7.3  & 0    & 11.1 & 0   & 0 \\
      & \VR  & 0    & 10.4 & \textbf{86.5} & 0    & 1.1  & 2.0  & 0   & 0 \\
      & \VG  & 0.4  & 0    & \textbf{66.7} & 26.6 & 1.0  & 5.3  & 0   & 0 \\
      & \VE  & 0    & 0    & 41.7 & 0    & 0    & 8.3  & \textbf{50.0} & 0 \\
      & \VMS & 0    & 0    & \textbf{66.7} & 33.3 & 0    & 0    & 0   & 0 \\
      & \CM  & 0    & 2.8  & \textbf{36.1} & 16.7 & 8.3  & 27.8 & 8.3 & 0 \\
    \midrule
    \multirow{7}{*}{\textbf{Qwen}}
      & \NV  & \textbf{74.8} & 17.3 & 0.6  & 0.9  & 0    & 4.7  & 0.2 & 1.3 \\
      & \VL  & 14.9 & \textbf{52.2} & 12.6 & 9.8  & 0    & 10.6 & 0   & 0 \\
      & \VR  & 0    & 13.4 & \textbf{84.8} & 0    & 0    & 0    & 0   & 1.7 \\
      & \VG  & 1.7  & 1.8  & \textbf{65.0} & 25.2 & 1.0  & 5.0  & 0   & 0.3 \\
      & \VE  & 0    & 41.7 & 0    & 0    & 0    & 0    & \textbf{58.3} & 0 \\
      & \VMS & 0    & 0    & \textbf{66.7} & 33.3 & 0    & 0    & 0   & 0 \\
      & \CM  & 0    & 18.1 & \textbf{30.6} & 8.3  & 8.3  & 18.1 & 16.7 & 0 \\
    \midrule
    \multirow{7}{*}{\textbf{Llama}}
      & \NV  & \textbf{70.8} & 20.2 & 1.4  & 1.4  & 0    & 6.0  & 0.1 & 0.1 \\
      & \VL  & 11.0 & \textbf{56.7} & 11.4 & 9.7  & 0    & 11.2 & 0   & 0 \\
      & \VR  & 0    & 12.9 & \textbf{85.1} & 0    & 0    & 2.0  & 0   & 0 \\
      & \VG  & 0.4  & 1.4  & \textbf{62.9} & 29.0 & 1.0  & 4.3  & 1.0 & 0 \\
      & \VE  & 0    & \textbf{75.0} & 0    & 16.7 & 0    & 8.3  & 0   & 0 \\
      & \VMS & 0    & 0    & \textbf{66.7} & 0    & 0    & 0    & 33.3 & 0 \\
      & \CM  & 0    & 18.1 & \textbf{30.6} & 8.3  & 8.3  & 18.1 & 16.7 & 0 \\
    \bottomrule
  \end{tabular}
  \caption{Variation transfer from source to target (\%, mean per document) per system, \MTbaseline\ condition, \iwslt. Rows: source variation category per system; columns: target variation category. Diagonal (preservation) in bold. Source distribution (ref., mean/doc): \NV: 74.5\% (88), \VL: 5.7\% (7), \VR: 8.9\% (10), \VG: 8.7\% (11), \VE: 0.4\% (4), \VMS: 0.3\% (1), \CM: 1.5\% (4).}
  \label{tab:var-transfer-baseline-IWSLT-simalign-output}
\end{table*}

\begin{table*}
  \centering
  \footnotesize
  \setlength{\tabcolsep}{4pt}
  \renewcommand{\arraystretch}{0.95}
  \begin{tabular}{ll|rrrrrrrr}
    \toprule
    System & Src & \NV & \VL & \VR & \VG & \VE & \VMS & \CM & $\emptyset$ \\
    \midrule
    \multirow{7}{*}{\textbf{Ref.}}
      & \NV  & \textbf{85.2} & 9.9  & 1.5  & 0.9  & 0.1  & 2.4  & 0.0 & 0.1 \\
      & \VL  & 7.1  & \textbf{63.0} & 18.7 & 4.1  & 1.8  & 4.1  & 0.4 & 0.9 \\
      & \VR  & 0.1  & 7.4  & \textbf{86.2} & 2.4  & 0.6  & 1.4  & 0.2 & 1.7 \\
      & \VG  & 13.2 & 13.6 & 9.6  & \textbf{49.9} & 5.4  & 3.9  & 3.1 & 1.1 \\
      & \VE  & 9.0  & 14.4 & 9.6  & 13.9 & \textbf{26.1} & 16.4 & 7.9 & 2.8 \\
      & \VMS & 12.3 & 24.1 & 17.6 & 0    & 10.2 & \textbf{35.8} & 0   & 0 \\
      & \CM  & 7.2  & \textbf{34.2} & 14.0 & 13.4 & 5.7  & 22.7 & 1.9 & 0.9 \\
    \midrule
    \multirow{7}{*}{\textbf{Euro22}}
      & \NV  & \textbf{93.0} & 3.3  & 1.6  & 0.5  & 0.1  & 1.3  & 0.0 & 0.0 \\
      & \VL  & 8.4  & \textbf{57.4} & 23.3 & 4.8  & 0.7  & 4.2  & 0   & 1.1 \\
      & \VR  & 0.2  & 8.7  & \textbf{85.0} & 2.3  & 0.2  & 2.0  & 0.1 & 1.6 \\
      & \VG  & 15.5 & 6.7  & 11.0 & \textbf{50.0} & 9.2  & 4.1  & 2.6 & 0.9 \\
      & \VE  & 9.7  & 16.2 & 2.7  & 23.2 & \textbf{34.0} & 9.3  & 2.1 & 2.8 \\
      & \VMS & 18.9 & 8.3  & 15.3 & 0    & 7.8  & \textbf{47.9} & 1.1 & 0.6 \\
      & \CM  & 9.1  & 30.3 & 9.7  & 8.7  & 6.4  & \textbf{31.2} & 3.8 & 0.9 \\
    \midrule
    \multirow{7}{*}{\textbf{Euro9}}
      & \NV  & \textbf{90.2} & 5.3  & 1.4  & 0.7  & 0.1  & 2.2  & 0   & 0.1 \\
      & \VL  & 7.7  & \textbf{60.0} & 21.8 & 3.6  & 2.0  & 3.1  & 0.1 & 1.7 \\
      & \VR  & 0.1  & 9.0  & \textbf{84.1} & 2.4  & 0.4  & 2.3  & 0.1 & 1.6 \\
      & \VG  & 14.8 & 5.7  & 9.9  & \textbf{50.9} & 9.8  & 3.2  & 4.4 & 1.3 \\
      & \VE  & 11.3 & 13.3 & 11.7 & 26.6 & \textbf{27.1} & 6.9  & 0   & 3.0 \\
      & \VMS & 19.5 & 14.6 & 13.8 & 0.6  & 7.2  & \textbf{41.7} & 0.6 & 2.1 \\
      & \CM  & 7.9  & 25.8 & 19.5 & 7.8  & 7.3  & \textbf{28.3} & 2.5 & 1.0 \\
    \midrule
    \multirow{7}{*}{\textbf{Qwen}}
      & \NV  & \textbf{93.5} & 3.2  & 1.5  & 0.5  & 0.1  & 1.1  & 0   & 0.1 \\
      & \VL  & 10.5 & \textbf{53.5} & 25.2 & 4.6  & 1.5  & 3.3  & 0   & 1.3 \\
      & \VR  & 0.7  & 7.3  & \textbf{85.1} & 2.2  & 0.3  & 2.7  & 0   & 1.7 \\
      & \VG  & 18.2 & 8.5  & 9.8  & \textbf{49.1} & 9.3  & 2.6  & 1.8 & 0.7 \\
      & \VE  & 6.0  & 16.5 & 2.1  & 17.0 & \textbf{44.4} & 10.7 & 0.6 & 2.8 \\
      & \VMS & 24.1 & 10.8 & 14.2 & 0    & 12.9 & \textbf{36.9} & 0.6 & 0.6 \\
      & \CM  & 12.9 & \textbf{26.2} & 11.7 & 12.8 & 7.2  & 25.9 & 2.5 & 0.9 \\
    \midrule
    \multirow{7}{*}{\textbf{Llama}}
      & \NV  & \textbf{93.2} & 3.4  & 1.4  & 0.7  & 0.1  & 1.1  & 0   & 0.1 \\
      & \VL  & 8.8  & \textbf{54.7} & 23.4 & 5.8  & 1.7  & 4.2  & 0   & 1.4 \\
      & \VR  & 0.6  & 7.3  & \textbf{86.0} & 2.4  & 0.3  & 1.7  & 0   & 1.7 \\
      & \VG  & 15.4 & 6.1  & 12.2 & \textbf{54.4} & 6.4  & 1.4  & 3.4 & 0.7 \\
      & \VE  & 12.2 & 10.9 & 2.7  & 31.9 & \textbf{36.5} & 3.0  & 0   & 2.8 \\
      & \VMS & 25.9 & 19.7 & 10.4 & 0    & 9.5  & \textbf{33.9} & 0   & 0.6 \\
      & \CM  & 9.1  & \textbf{32.1} & 10.8 & 8.7  & 8.2  & 26.9 & 3.3 & 0.9 \\
    \bottomrule
  \end{tabular}
  \caption{Variation transfer from source to target (\%, mean per document) per system, \MTmoslem\ condition, \ACL. Rows: source variation category per system; columns: target variation category. Diagonal (preservation) in bold. Source distribution (ref., mean/doc): \NV: 72.3\% (198), \VL: 7.9\% (22), \VR: 9.0\% (25), \VG: 7.7\% (21), \VE: 1.1\% (4), \VMS: 0.6\% (2), \CM: 1.3\% (5).}
  \label{tab:var-transfer-moslem-ACL-simalign-output}
\end{table*}

\begin{table*}
  \centering
  \footnotesize
  \setlength{\tabcolsep}{4pt}
  \renewcommand{\arraystretch}{0.95}
  \begin{tabular}{ll|rrrrrrrr}
    \toprule
    System & Src & \NV & \VL & \VR & \VG & \VE & \VMS & \CM & $\emptyset$ \\
    \midrule
    \multirow{7}{*}{\textbf{Ref.}}
      & \NV  & \textbf{72.6} & 21.1 & 1.4  & 0.8  & 0    & 4.0  & 0   & 0.1 \\
      & \VL  & 3.5  & \textbf{77.2} & 8.1  & 0.6  & 0    & 10.6 & 0   & 0 \\
      & \VR  & 0    & 11.5 & \textbf{86.4} & 0    & 0    & 1.0  & 1.1 & 0 \\
      & \VG  & 0    & 0    & \textbf{78.6} & 14.2 & 2.0  & 5.3  & 0   & 0 \\
      & \VE  & 0    & 41.7 & 0    & 0    & \textbf{58.3} & 0    & 0   & 0 \\
      & \VMS & 0    & 0    & \textbf{66.7} & 0    & 0    & 33.3 & 0   & 0 \\
      & \CM  & 0    & 8.3  & 20.8 & 8.3  & 8.3  & \textbf{45.8} & 8.3 & 0 \\
    \midrule
    \multirow{7}{*}{\textbf{Euro22}}
      & \NV  & \textbf{86.7} & 7.5  & 1.4  & 0.7  & 0.1  & 3.5  & 0.1 & 0 \\
      & \VL  & 21.6 & \textbf{40.3} & 12.6 & 8.5  & 0    & 16.9 & 0   & 0 \\
      & \VR  & 1.2  & 10.4 & \textbf{84.1} & 0    & 2.2  & 2.0  & 0   & 0 \\
      & \VG  & 12.1 & 0    & \textbf{65.3} & 21.2 & 1.4  & 0    & 0   & 0 \\
      & \VE  & 0    & \textbf{91.7} & 0    & 0    & 8.3  & 0    & 0   & 0 \\
      & \VMS & 0    & 0    & \textbf{100.0} & 0   & 0    & 0    & 0   & 0 \\
      & \CM  & 8.3  & 19.4 & \textbf{36.1} & 8.3  & 8.3  & 11.1 & 8.3 & 0 \\
    \midrule
    \multirow{7}{*}{\textbf{Euro9}}
      & \NV  & \textbf{80.4} & 12.8 & 1.1  & 0.4  & 0    & 5.0  & 0.1 & 0.1 \\
      & \VL  & 16.6 & \textbf{52.3} & 11.4 & 7.8  & 0    & 11.8 & 0   & 0 \\
      & \VR  & 0    & 10.3 & \textbf{86.6} & 0    & 1.1  & 2.0  & 0   & 0 \\
      & \VG  & 8.5  & 0    & \textbf{63.9} & 26.6 & 1.0  & 0    & 0   & 0 \\
      & \VE  & 0    & 0    & 41.7 & 8.3  & 0    & 0    & \textbf{50.0} & 0 \\
      & \VMS & 0    & \textbf{33.3} & \textbf{33.3} & \textbf{33.3} & 0 & 0 & 0 & 0 \\
      & \CM  & 8.3  & 20.8 & \textbf{36.1} & 8.3  & 8.3  & 9.7  & 8.3 & 0 \\
    \midrule
    \multirow{7}{*}{\textbf{Qwen}}
      & \NV  & \textbf{89.8} & 5.2  & 1.1  & 0.7  & 0    & 2.7  & 0.1 & 0.4 \\
      & \VL  & 22.1 & \textbf{43.1} & 9.3  & 6.0  & 0    & 13.1 & 0   & 6.4 \\
      & \VR  & 3.0  & 9.2  & \textbf{85.6} & 0    & 1.1  & 1.1  & 0   & 0 \\
      & \VG  & 8.4  & 0    & \textbf{65.4} & 23.2 & 1.0  & 2.0  & 0   & 0 \\
      & \VE  & 8.3  & 25.0 & 0    & 8.3  & \textbf{58.3} & 0    & 0   & 0 \\
      & \VMS & 0    & 0    & \textbf{33.3} & \textbf{33.3} & 0 & 0 & 0 & \textbf{33.3} \\
      & \CM  & 8.3  & 2.8  & \textbf{36.1} & 8.3  & 8.3  & 11.1 & 25.0 & 0 \\
    \midrule
    \multirow{7}{*}{\textbf{Llama}}
      & \NV  & \textbf{88.9} & 6.6  & 1.5  & 0.7  & 0.1  & 2.2  & 0   & 0 \\
      & \VL  & 26.2 & \textbf{40.7} & 10.5 & 4.8  & 0    & 17.8 & 0   & 0 \\
      & \VR  & 1.2  & 9.8  & \textbf{86.6} & 0    & 1.1  & 1.2  & 0   & 0 \\
      & \VG  & 13.7 & 0    & \textbf{58.7} & 26.6 & 1.0  & 0    & 0   & 0 \\
      & \VE  & 0    & 25.0 & 0    & 16.7 & \textbf{58.3} & 0    & 0   & 0 \\
      & \VMS & 0    & 0    & \textbf{66.7} & 0    & 0    & 0    & 33.3 & 0 \\
      & \CM  & 8.3  & 0    & \textbf{38.9} & 0    & 8.3  & 36.1 & 8.3 & 0 \\
    \bottomrule
  \end{tabular}
  \caption{Variation transfer from source to target (\%, mean per document) per system, \MTmoslem\ condition, \iwslt. Rows: source variation category per system; columns: target variation category. Diagonal (preservation) in bold. Source distribution (ref., mean/doc): \NV: 74.5\% (88), \VL: 5.7\% (7), \VR: 8.9\% (10), \VG: 8.7\% (11), \VE: 0.4\% (4), \VMS: 0.3\% (1), \CM: 1.5\% (4).}
  \label{tab:var-transfer-moslem-IWSLT-simalign-output}
\end{table*}

\subsection{Legitimate Variation Example: \textit{reported speech}}
\label{app:reported-speech-variation-extended}

Table~\ref{tab:reported-speech-variation} shows translations of the sentence below by all systems. Within a single sentence, the reference alternates between \textit{parole rapportée} and \textit{discours rapporté} as renderings of \textit{reported speech}, a stylistic choice that a consistency metric would penalize as inconsistency, although both forms are standard French equivalents of the same linguistic phenomenon. All four \MTbaseline\ MT systems instead produce a single form repeated.

\begin{table*}[h!]
\centering\small
\renewcommand{\arraystretch}{1.1}
\begin{tabular}{@{}lp{0.82\textwidth}@{}}
\toprule
\textbf{Source} & We dealt with time-stamped events governed by a \textbf{reported speech} verb or otherwise appearing in \textbf{reported speech}. \\
\midrule
\textbf{Reference} & Nous avons traité des événements datés régis par un verbe de \textbf{parole rapportée} ou apparaissant dans un \textbf{discours rapporté}. \\
\addlinespace
\textbf{Llama} & Nous avons traité des événements datés régis par un verbe de \textbf{discours rapporté} ou apparaissant autrement dans le \textbf{discours rapporté}. \\
\textbf{Euro22}   & Nous avons traité des événements marqués par un verbe de \textbf{discours rapporté} ou apparaissant autrement dans un \textbf{discours rapporté}. \\
\textbf{Euro9}    & Nous avons traité des événements marqués par un horodatage régis par un verbe de \textbf{discours rapporté} ou apparaissant autrement dans un \textbf{discours rapporté}. \\
\textbf{Qwen}  & Nous avons traité des événements datés de temps, régis par un verbe de \textbf{discours direct} ou autrement apparaissant dans un \textbf{discours rapporté}. \\
\bottomrule
\end{tabular}
\caption{A sentence from the \ACL corpus where the reference alternates between \textit{parole rapportée} and \textit{discours rapporté} for the same source term \textit{reported speech}, both standard French renderings of the same concept. The four MT systems produce a single repeated form (Qwen substitutes \textit{discours direct}, a different concept).} %(ACL\_13)
\label{tab:reported-speech-variation}
\end{table*}

\subsection{Legitimate Variation Example: \textit{pitch}}
\label{app:pitch-consistency}

Table~\ref{tab:pitch-consistency} shows three of the 14 occurrences of \textit{pitch} (glossary: \textit{tonie}) in a single \ACL document, translated as either \textit{hauteur} (for pitch values) or \textit{m\'elodique} (in compound expressions such as \textit{contour m\'elodique} or \textit{patron m\'elodique}). All renderings are contextually appropriate, yet a consistency metric severely penalizes the alternation.

\subsection{Variation Type Variability Example: \textit{reinforcement learning}}
\label{app:rl-ctv-example}

Table~\ref{tab:rl-ctv-example} shows three occurrences of the acronym \textit{RL} (a \VG variant of \textit{reinforcement learning}) in a single \ACL document. The acronym is preserved in one context but expanded into different forms in others, illustrating the within-document variability of \VG transfer.

\begin{table*}
\centering\small
\setlength{\tabcolsep}{2pt}
\renewcommand{\arraystretch}{0.9}
\begin{tabular}{@{}p{\columnwidth}p{\columnwidth}@{}}
\toprule
\textbf{Source} & \textbf{Reference} \\
\midrule
\ldots via Reinforcement Learning (\textbf{RL}). & \ldots via l'apprentissage par renforcement (\textbf{RL}). \hfill \textit{\VG$\to$\VG} \cmark \\
\ldots a data-driven method such as \textbf{RL}. & \ldots des m\'ethodes d'apprentissage statistique telles que \textbf{l'apprentissage par renforcement}. \hfill \textit{\VG$\to$\NV} \xmark \\
An \textbf{RL} agent compares different \ldots & Un agent \textbf{apprenant par renforcement} compare diff\'erentes \ldots \hfill \textit{\VG$\to$\CM} \xmark \\
\bottomrule
\end{tabular}
\caption{Cross-term variation for \textit{reinforcement learning $\to$ apprentissage par renforcement}: \textit{RL} is preserved or expanded, illustrating type-dependent variability.}
\label{tab:rl-ctv-example}
\end{table*}

\begin{table}[H]
\centering\small
\setlength{\tabcolsep}{2pt}
\renewcommand{\arraystretch}{0.9}
\begin{tabular}{@{}p{0.47\columnwidth}p{0.47\columnwidth}@{}}
\toprule
\textbf{Source} & \textbf{Reference} \\
\midrule
\ldots \textbf{pitch} values were extracted and averaged\ldots & \ldots les valeurs de \textbf{hauteur} ont \'{e}t\'{e} extraites\ldots \\
\ldots more \textbf{pitch} movement on penultimate syllables\ldots & \ldots plus de mouvement \textbf{m\'{e}lodique} sur la p\'{e}nulti\`{e}me\ldots \\
\ldots a specific \textbf{pitch} pattern on the penultimate syllable. & \ldots un sch\'{e}ma \textbf{m\'{e}lodique} sp\'{e}cifique sur la syllabe p\'{e}nulti\`{e}me. \\
\bottomrule
\end{tabular}
\caption{The term \textit{pitch} (glossary: \textit{tonie}) translated as 2 distinct lexical variants (\textit{hauteur}, \textit{m\'elodique}) across 14 occurrences by the Reference. All translations are contextually appropriate, yet a consistency metric results in a low score for this term.}
\label{tab:pitch-consistency}
\end{table}

\subsection{Unevaluated Terms}
\label{app:unevaluated}

Tables~\ref{tab:never_evaluated} and~\ref{tab:never_evaluated_iwslt} characterize the terms that remain unevaluated by any of the three dimensions. For \ACL, 3.2\% of occurrences (12.1\% of distinct terms) are never evaluated; for \iwslt, 2.5\% (8.1\%). These residual terms are hapax occurrences absent from the glossary and without a \CTV-eligible relationship to a preferred term, representing a small but structurally irreducible gap.

\begin{table*}
\centering
\caption{Average proportion of terms covered by no evaluation method per document (\textit{occ} = occurrences; \textit{dist} = distinct terms; never evaluated $\equiv$ not in glossary $\wedge$ hapax ($k{=}1$) $\wedge$ not ITC-eligible)}
\label{tab:never_evaluated}
\footnotesize
\setlength{\tabcolsep}{4pt}
\begin{tabular}{lrrrc}
\toprule
Category & \multicolumn{2}{c}{Avg occ/doc} & \multicolumn{2}{c}{Avg dist/doc} \\
\cmidrule(lr){2-3}\cmidrule(lr){4-5}
& Avg $N$ & \% & Avg $N$ & \% \\
\midrule
All terms & 273.9 & --- & 68.7 & --- \\
\midrule
Eval.\ by $\geq$1 method & 265.5 & 96.8\% & 60.3 & 87.9\% \\
\midrule
\textbf{Never evaluated} & \textbf{8.4} & \textbf{3.2\%} & \textbf{8.4} & \textbf{12.1\%} \\
\midrule
\quad Hapax ($k{=}1$, excl.\ by TLC) & 34.2 & 12.5\% & 34.2 & 49.8\% \\
\quad Not in glossary (excl.\ by EM) & 29.5 & 10.8\% & 6.7 & 9.8\% \\
\midrule
\multicolumn{5}{l}{\footnotesize Averaged over 31\,documents. Never evaluated $\equiv$ not in glossary $\wedge$ hapax $\wedge$ not ITC-eligible.} \\
\bottomrule
\end{tabular}
\end{table*}
\begin{table*}
\centering
\footnotesize
\setlength{\tabcolsep}{4pt}
\begin{tabular}{lrrrr}
\toprule
Category & \multicolumn{2}{c}{Avg occ/doc} & \multicolumn{2}{c}{Avg dist/doc} \\
\cmidrule(lr){2-3}\cmidrule(lr){4-5}
& Avg $N$ & \% & Avg $N$ & \% \\
\midrule
All terms & 120.3 & --- & 35.2 & --- \\
\midrule
Eval.\ by $\geq$1 method & 117.3 & 97.5\% & 32.2 & 91.9\% \\
\midrule
\textbf{Never evaluated} & \textbf{3.0} & \textbf{2.5\%} & \textbf{3.0} & \textbf{8.1\%} \\
\midrule
\quad Hapax ($k{=}1$, excl.\ by TLC) & 16.5 & 13.7\% & 16.5 & 46.9\% \\
\quad Not in glossary (excl.\ by EM) & 7.4 & 6.2\% & 2.2 & 6.2\% \\
\midrule
\multicolumn{5}{l}{\footnotesize Averaged over 10\,documents. Never evaluated $\equiv$ not in glossary $\wedge$ hapax $\wedge$ not ITC-eligible.} \\
\bottomrule
\end{tabular}
\caption{Average proportion of terms covered by no evaluation method per document (\textit{occ} = occurrences; \textit{dist} = distinct terms; never evaluated $\equiv$ not in glossary $\wedge$ hapax ($k{=}1$) $\wedge$ not ITC-eligible)}
\label{tab:never_evaluated_iwslt}
\end{table*}

\subsection{TCR Metric Divergence Example}
\label{app:tlc-example}

Table~\ref{tab:tlc-example} illustrates how the three \ITC metrics diverge on the same data. For 5 occurrences of \textit{neural network} (glossary: \textit{réseau de neurones}), \ITC-EM scores 0.40, \ITC-PR 0.60, and \ITC-T 0.50.

\begin{table}[H]
\centering
\small
\begin{tabular}{@{}clccc@{}}
\toprule
& \textbf{Translation} & \textbf{EM} & \textbf{PR} & \textbf{Temporal} \\
\midrule
$o_1$ & réseau neuronal & \texttimes & $=$   & \\
$o_2$ & réseau neuronal & \texttimes & $=$   & $=$ \\
$o_3$ & réseau de neurones & \checkmark & $\neq$ & $\neq$ \\
$o_4$ & réseau de neurones & \checkmark & $\neq$ & $=$ \\
$o_5$ & réseau neuronal & \texttimes & $=$   & $\neq$ \\
\midrule
\multicolumn{2}{@{}l}{Score} & 2/5 & 3/5 & 2/4 \\
\bottomrule
\end{tabular}
\caption{Translations of \textit{neural network} (glossary: \textit{réseau de neurones}). \textbf{EM}: glossary match; \textbf{PR}: agrees with pseudo-ref ($o_1$); \textbf{Temporal}: agrees with previous occ.}
\label{tab:tlc-example}
\end{table}

\subsection{Scores Breakdown by Variation Label}
\label{app:scores-breakdown}

\subsubsection{Baseline Condition}

\paragraph{Term-level accuracy (EM).}
Table~\ref{tab:var-baseline-gcem-micro-restr} reports EM Micro broken down by source variation category. EM ranges from 70.0--74.1 on \iwslt and 77.8--81.5 on \ACL, with Qwen leading on \iwslt (74.05) and Ref on \ACL (81.52). \NV terms achieve 72--85\% EM accuracy across systems. Performance drops sharply for variants: \VL reaches only 34--55\%, \VG 42--74\%, and \VMS and \CM score 0\% across all systems on both corpora. Since \NV accounts for 72--74\% of occurrences, the aggregate EM is structurally dominated by preferred-term performance.

\paragraph{Translation consistency (\ITC).}
Tables~\ref{tab:var-baseline-tlc-canpseudo-ref-first-weighted},~\ref{tab:var-baseline-tlc-cansequential-weighted}, and~\ref{tab:var-baseline-tlc-canem-macro-weighted} report the three \ITC variants broken down by variation category. Under the least restrictive variant (\ITC-PR), the reference dominates on \iwslt (95.09 vs.\ 84.4--88.1 for MT systems) but differences narrow on \ACL (86.8--89.3). The per-variation pattern is consistent across metrics: \NV terms score substantially higher (88--95\%) than \VL (47--55\% on \iwslt, 78--88\% on \ACL) or \VG (53--89\% depending on corpus and system). \CM scores are low but non-zero (6--24\%), indicating that the few compound variants with $k \geq 2$ occurrences are rarely translated consistently.

A notable observation concerns system ranking stability. On \ACL under baseline, each \ITC variant ranks a different system as top-1: \ITC-EM favors the reference (82.99), \ITC-PR favors Llama (89.31), and \ITC-T favors E9 (87.93). No single system dominates consistency across all operationalizations, indicating that the choice of consistency metric is not neutral and can substantially affect system ranking.

\paragraph{Cross-term variation (\CTV).}
Table~\ref{tab:var-baseline-itc-coherence} reports variation preservation by source variation category. On \iwslt, the reference achieves the highest \CTV (86.67), outperforming all MT systems (68--86\%). On \ACL, the pattern reverses: the reference scores lowest (66.06) while Llama leads (74.63). This asymmetry is explained by the target-side amplification documented in Section~\ref{ssec:variation-transfer}.

Per variation type, \VR preservation is consistently high across all systems (82--100\% on \iwslt, 81--87\% on \ACL), confirming that reductions transfer reliably. \VG coherence varies more: the reference achieves 90\% on \iwslt but only 58\% on \ACL, while MT systems range from 50--86\% on \iwslt and 42--78\% on \ACL. \VL coherence shows the widest system-dependent spread (50--86\% on \iwslt, 57--72\% on \ACL), and \VE is both rare and variable.

\subsubsection{Glossary-Constrained Condition (\MTmoslem)}

\paragraph{Term-level accuracy (EM).}
Under \MTmoslem (Table~\ref{tab:var-prompts-with-glossary-moslem-gcem-micro-restr}), EM Micro increases substantially for all MT systems: +10--17 points on \iwslt (from 70--74\% to 78--90\%) and +6--10 points on \ACL (from 78--80\% to 85--89\%), confirming that baseline systems already possess the correct terms but do not always produce them without guidance. The improvement is concentrated on \NV terms (+14--18 points on \iwslt, +8--12 points on \ACL), with smaller gains on \VR and \VG. \VMS and \CM remain at 0\%.

\paragraph{Translation consistency (\ITC).}
\ITC metrics also improve under \MTmoslem (Tables~\ref{tab:var-prompts-with-glossary-moslem-tlc-canpseudo-ref-first-weighted},~\ref{tab:var-prompts-with-glossary-moslem-tlc-cansequential-weighted},~\ref{tab:var-prompts-with-glossary-moslem-tlc-canem-macro-weighted}). On \NV terms, \ITC-PR rises from 86--88\% to 91--95\% on \iwslt and from 88--90\% to 93--95\% on \ACL, indicating that forcing preferred-term usage reduces translation inconsistency. The improvement is smaller for variant categories, which are not directly targeted by the glossary constraint.

\paragraph{Cross-term variation (\CTV).}
Under \MTmoslem (Table~\ref{tab:var-prompts-with-glossary-moslem-itc-coherence}), \CTV drops for most MT systems compared to baseline, particularly on \ACL (from 72--75\% to 67--69\%). On \iwslt, \VG coherence improves slightly for some systems (Qwen: 50\% $\to$ 90\%), but \VR coherence decreases (95--100\% $\to$ 83--96\%). This confirms the tension identified in the main analysis: the glossary constraint improves accuracy and consistency at the cost of suppressing the target-side variation that \CTV measures. The reference, unchanged across conditions, serves as a stable baseline.

\begin{table*}
    \centering
    \resizebox{\textwidth}{!}{
    \begin{tabular}{lrrrrrrr|rrrrrrr}
    \toprule
    Var. & \multicolumn{7}{c|}{\ACL} & \multicolumn{7}{c}{\iwslt} \\
    \cmidrule(lr){2-8} \cmidrule(lr){9-15}
    & $N$ & \% & Ref$_S$ & Euro22 & Euro9 & Qwen & Llama& $N$ & \% & Ref & Euro22 & Euro9 & Qwen & Llama  \\
    \midrule
    \NV & 197.7 & 72.2\% & \textbf{85.31} & 83.15 & 83.77 & 81.54 & 81.80& 88.4 & 73.5\% & 72.63 & 72.47 & 73.07 & \textbf{74.87} & 70.85  \\
    \VR & 24.6 & 9.0\% & 87.09 & 85.13 & 82.00 & 86.15 & \textbf{88.11}& 9.9 & 8.2\% & 67.92 & 65.42 & 65.42 & \textbf{68.75} & 66.67  \\
    \VMS & 1.7 & 0.6\% & \textbf{0.00} & \textbf{0.00} & \textbf{0.00} & \textbf{0.00} & \textbf{0.00}& 0.3 & 0.2\% & \textbf{0.00} & \textbf{0.00} & \textbf{0.00} & \textbf{0.00} & \textbf{0.00}  \\
    \VE & 3.2 & 1.2\% & 6.45 & 6.45 & \textbf{23.12} & 2.15 & \textbf{23.12}& 0.7 & 0.6\% & \textbf{0.00} & \textbf{0.00} & \textbf{0.00} & \textbf{0.00} & \textbf{0.00}  \\
    \VG & 20.6 & 7.5\% & \textbf{45.76} & 43.11 & 42.51 & 45.08 & 42.00& 11.4 & 9.5\% & \textbf{73.78} & 58.15 & 59.26 & 65.87 & 56.35  \\
    \VL & 22.0 & 8.0\% & 40.35 & 47.79 & \textbf{49.63} & 38.59 & 45.24& 7.4 & 6.2\% & 34.66 & 45.98 & \textbf{55.24} & 45.77 & 44.25  \\
    \CM & 4.0 & 1.5\% & \textbf{0.00} & \textbf{0.00} & \textbf{0.00} & \textbf{0.00} & \textbf{0.00}& 2.2 & 1.8\% & \textbf{0.00} & \textbf{0.00} & \textbf{0.00} & \textbf{0.00} & \textbf{0.00}  \\
    \bottomrule
    \end{tabular}
    }
    \caption{EM accuracy (\%) by source variation category under the \MTbaseline condition. $N$: avg.\ occurrences per document; \%: share of total occurrences. Best system per corpus in bold.}
    \label{tab:var-baseline-gcem-micro-restr}
\end{table*}

\begin{table*}[!ht]
    \centering
    \resizebox{\textwidth}{!}{
    \begin{tabular}{lrrrrrrr|rrrrrrr}
    \toprule
    Var. & \multicolumn{7}{c|}{\ACL} & \multicolumn{7}{c}{\iwslt} \\
    \cmidrule(lr){2-8} \cmidrule(lr){9-15}
    & $N$ & \% & Ref$_S$ & Euro22 & Euro9 & Qwen & Llama& $N$ & \% & Ref & Euro22 & Euro9 & Qwen & Llama  \\
    \midrule
    \NV & 22.4 & 64.9\% & 88.96 & \textbf{90.82} & 90.67 & 88.20 & 90.11& 13.2 & 70.6\% & \textbf{95.41} & 85.58 & 88.26 & 88.35 & 87.71  \\
    \VR & 3.8 & 11.0\% & 89.95 & 89.62 & 84.55 & 88.54 & \textbf{91.99}& 2.4 & 12.8\% & \textbf{94.24} & 84.86 & 82.94 & 82.38 & 87.44  \\
    \VMS & 0.0 & 0.1\% & \textbf{1.61} & \textbf{1.61} & \textbf{1.61} & \textbf{1.61} & \textbf{1.61}& 0.0 & 0.0\% & \textbf{0.00} & \textbf{0.00} & \textbf{0.00} & \textbf{0.00} & \textbf{0.00}  \\
    \VE & 0.5 & 1.3\% & \textbf{26.87} & 22.65 & 25.84 & 18.62 & 24.48& 0.1 & 0.5\% & 8.00 & 6.00 & \textbf{10.00} & \textbf{10.00} & 6.00  \\
    \VG & 3.5 & 10.0\% & 55.89 & 65.18 & \textbf{67.74} & 65.19 & 67.68& 1.3 & 7.0\% & \textbf{88.86} & 65.36 & 73.71 & 80.91 & 58.38  \\
    \VL & 3.9 & 11.3\% & \textbf{87.66} & 78.66 & 85.06 & 81.45 & 84.19& 1.3 & 7.0\% & 52.95 & 52.50 & 49.92 & \textbf{53.41} & 49.47  \\
    \CM & 0.5 & 1.3\% & 19.97 & 20.89 & 21.89 & 21.12 & \textbf{24.19}& 0.4 & 2.1\% & \textbf{19.17} & 12.50 & 12.50 & 14.17 & 14.17  \\
    \bottomrule
    \end{tabular}
    }
    \caption{\ITC pseudo-ref (\%) by source variation category under the \MTbaseline condition. $N$: avg.\ distinct terms per document with $k \geq 2$ occurrences; \%: share of total. Best system per corpus in bold.}
    \label{tab:var-baseline-tlc-canpseudo-ref-first-weighted}
\end{table*}

\begin{table*}[!ht]
    \centering
    \resizebox{\textwidth}{!}{
    \begin{tabular}{lrrrrrrr|rrrrrrr}
    \toprule
    Var. & \multicolumn{7}{c|}{\ACL} & \multicolumn{7}{c}{\iwslt} \\
    \cmidrule(lr){2-8} \cmidrule(lr){9-15}
    & $N$ & \% & Ref$_S$ & Euro22 & Euro9 & Qwen & Llama& $N$ & \% & Ref & Euro22 & Euro9 & Qwen & Llama  \\
    \midrule
    \NV & 22.4 & 64.9\% & 87.98 & \textbf{90.79} & 89.69 & 87.50 & 88.71& 13.2 & 70.6\% & \textbf{94.81} & 86.60 & 86.63 & 85.78 & 84.85  \\
    \VR & 3.8 & 11.0\% & 84.97 & 86.31 & 82.64 & 84.23 & \textbf{89.23}& 2.4 & 12.8\% & \textbf{89.43} & 72.43 & 72.29 & 70.84 & 80.71  \\
    \VMS & 0.0 & 0.1\% & \textbf{0.00} & \textbf{0.00} & \textbf{0.00} & \textbf{0.00} & \textbf{0.00}& 0.0 & 0.0\% & \textbf{0.00} & \textbf{0.00} & \textbf{0.00} & \textbf{0.00} & \textbf{0.00}  \\
    \VE & 0.5 & 1.3\% & \textbf{24.20} & 18.00 & 20.83 & 13.86 & 24.13& 0.1 & 0.5\% & 5.00 & 0.00 & \textbf{10.00} & \textbf{10.00} & 5.00  \\
    \VG & 3.5 & 10.0\% & 54.49 & 60.30 & 63.54 & 64.16 & \textbf{67.23}& 1.3 & 7.0\% & \textbf{88.24} & 55.36 & 61.23 & 73.32 & 53.82  \\
    \VL & 3.9 & 11.3\% & \textbf{84.00} & 74.87 & 81.65 & 78.19 & 77.61& 1.3 & 7.0\% & \textbf{51.73} & 47.07 & 46.61 & 49.64 & 44.53  \\
    \CM & 0.5 & 1.3\% & 13.30 & 14.29 & 17.19 & 15.44 & \textbf{19.88}& 0.4 & 2.1\% & \textbf{17.78} & 6.67 & 6.67 & 8.89 & 8.89  \\
    \bottomrule
    \end{tabular}
    }
    \caption{\ITC temporal (\%) by source variation category under the \MTbaseline condition. $N$: avg.\ distinct terms per document with $k \geq 2$ occurrences; \%: share of total. Best system per corpus in bold.}
    \label{tab:var-baseline-tlc-cansequential-weighted}
\end{table*}

\begin{table*}[!ht]
    \centering
    \resizebox{\textwidth}{!}{
    \begin{tabular}{lrrrrrrr|rrrrrrr}
    \toprule
    Var. & \multicolumn{7}{c|}{\ACL} & \multicolumn{7}{c}{\iwslt} \\
    \cmidrule(lr){2-8} \cmidrule(lr){9-15}
    & $N$ & \% & Ref$_S$ & Euro22 & Euro9 & Qwen & Llama& $N$ & \% & Ref & Euro22 & Euro9 & Qwen & Llama  \\
    \midrule
    \NV & 22.4 & 80.7\% & \textbf{86.57} & 84.57 & 85.17 & 82.95 & 83.33& 13.2 & 80.0\% & 74.01 & 73.92 & 74.05 & \textbf{76.70} & 72.28  \\
    \VR & 1.4 & 5.1\% & 87.82 & 86.36 & 83.19 & 88.58 & \textbf{89.21}& 1.2 & 7.3\% & 72.08 & 69.58 & 69.58 & \textbf{75.42} & 70.83  \\
    \VMS & 0.0 & 0.1\% & \textbf{0.00} & \textbf{0.00} & \textbf{0.00} & \textbf{0.00} & \textbf{0.00}& 0.0 & 0.0\% & \textbf{0.00} & \textbf{0.00} & \textbf{0.00} & \textbf{0.00} & \textbf{0.00}  \\
    \VE & 0.2 & 0.7\% & 1.61 & 1.61 & \textbf{7.53} & 0.54 & \textbf{7.53}& 0.1 & 0.6\% & \textbf{0.00} & \textbf{0.00} & \textbf{0.00} & \textbf{0.00} & \textbf{0.00}  \\
    \VG & 2.3 & 8.1\% & \textbf{44.78} & 41.81 & 41.06 & 44.45 & 40.71& 1.3 & 7.9\% & \textbf{76.61} & 60.36 & 61.46 & 68.66 & 58.49  \\
    \VL & 1.4 & 5.0\% & 37.90 & 46.82 & \textbf{47.36} & 38.14 & 42.47& 0.7 & 4.2\% & 13.89 & \textbf{25.35} & 25.03 & 25.00 & 23.76  \\
    \CM & 0.1 & 0.2\% & \textbf{0.00} & \textbf{0.00} & \textbf{0.00} & \textbf{0.00} & \textbf{0.00}& 0.0 & 0.0\% & \textbf{0.00} & \textbf{0.00} & \textbf{0.00} & \textbf{0.00} & \textbf{0.00}  \\
    \bottomrule
    \end{tabular}
    }
    \caption{\ITC$_{\text{EM}_{\text{macro}}}$ (\%) by source variation category under the \MTbaseline condition. $N$: avg.\ in-glossary distinct terms per document with $k \geq 2$ occurrences; \%: share of total. Best system per corpus in bold.}
    \label{tab:var-baseline-tlc-canem-macro-weighted}
\end{table*}

\begin{table*}[!ht]
    \centering
    \resizebox{\textwidth}{!}{
    \begin{tabular}{lrrrrrrr|rrrrrrr}
    \toprule
    Var. & \multicolumn{7}{c|}{\ACL} & \multicolumn{7}{c}{\iwslt} \\
    \cmidrule(lr){2-8} \cmidrule(lr){9-15}
    & $N$ & \% & Ref$_S$ & Euro22 & Euro9 & Qwen & Llama& $N$ & \% & Ref & Euro22 & Euro9 & Qwen & Llama  \\
    \midrule
    \VG & 9.1 & 34.2\% & 57.95 & 69.92 & 72.06 & 67.73 & \textbf{77.83}& 1.0 & 33.3\% & \textbf{90.00} & 85.71 & 85.71 & 50.00 & 85.71  \\
    \VR & 8.4 & 31.5\% & \textbf{87.36} & 84.85 & 81.03 & 84.65 & 83.95& 1.1 & 36.7\% & 81.82 & 95.65 & \textbf{100.00} & \textbf{100.00} & \textbf{100.00}  \\
    \VE & 1.3 & 4.8\% & \textbf{40.00} & 32.43 & 29.73 & 23.68 & 28.21& 0.2 & 6.7\% & \textbf{100.00} & \textbf{100.00} & 50.00 & 0.00 & 0.00  \\
    \VL & 7.9 & 29.5\% & 56.97 & 70.35 & \textbf{72.12} & 70.76 & 69.71& 0.7 & 23.3\% & \textbf{85.71} & 60.00 & 60.00 & 50.00 & 60.00  \\
    \bottomrule
    \end{tabular}
    }
    \caption{\CTV variation preservation (\%) by source variation category under the \MTbaseline condition. $N$: avg.\ eligible occurrences per document (variants whose head term is translated as \NV); \%: share of eligible total. Best system per corpus in bold.}
    \label{tab:var-baseline-itc-coherence}
\end{table*}

\begin{table*}
    \centering
    \resizebox{\textwidth}{!}{
    \begin{tabular}{lrrrrrrr|rrrrrrr}
    \toprule
    Var. & \multicolumn{7}{c|}{\ACL} & \multicolumn{7}{c}{\iwslt} \\
    \cmidrule(lr){2-8} \cmidrule(lr){9-15}
    & $N$ & \% & Ref$_S$ & Euro22 & Euro9 & Qwen & Llama& $N$ & \% & Ref & Euro22 & Euro9 & Qwen & Llama  \\
    \midrule
    \NV & 197.7 & 72.2\% & 85.31 & 93.23 & 90.38 & \textbf{93.96} & 93.63& 88.4 & 73.5\% & 72.63 & 87.12 & 80.69 & \textbf{90.22} & 89.31  \\
    \VR & 24.6 & 9.0\% & 87.09 & 88.75 & 86.14 & 89.39 & \textbf{89.56}& 9.9 & 8.2\% & 67.92 & 70.00 & 65.42 & \textbf{72.50} & \textbf{72.50}  \\
    \VMS & 1.7 & 0.6\% & \textbf{0.00} & \textbf{0.00} & \textbf{0.00} & \textbf{0.00} & \textbf{0.00}& 0.3 & 0.2\% & \textbf{0.00} & \textbf{0.00} & \textbf{0.00} & \textbf{0.00} & \textbf{0.00}  \\
    \VE & 3.2 & 1.2\% & 6.45 & 12.90 & 12.90 & \textbf{24.73} & 22.58& 0.7 & 0.6\% & \textbf{0.00} & \textbf{0.00} & \textbf{0.00} & \textbf{0.00} & \textbf{0.00}  \\
    \VG & 20.6 & 7.5\% & 45.76 & 44.40 & 40.98 & 43.41 & \textbf{48.61}& 11.4 & 9.5\% & 73.78 & 63.41 & 58.57 & \textbf{78.41} & 76.27  \\
    \VL & 22.0 & 8.0\% & 40.35 & 47.31 & \textbf{48.12} & 37.01 & 44.01& 7.4 & 6.2\% & 34.66 & 46.54 & \textbf{48.08} & 47.71 & 42.59  \\
    \CM & 4.0 & 1.5\% & \textbf{0.00} & \textbf{0.00} & \textbf{0.00} & \textbf{0.00} & \textbf{0.00}& 2.2 & 1.8\% & \textbf{0.00} & \textbf{0.00} & \textbf{0.00} & \textbf{0.00} & \textbf{0.00}  \\
    \bottomrule
    \end{tabular}
    }
    \caption{EM accuracy (\%) by source variation category under the \MTmoslem condition. $N$: avg.\ occurrences per document; \%: share of total occurrences. Best system per corpus in bold.}
    \label{tab:var-prompts-with-glossary-moslem-gcem-micro-restr}
\end{table*}

\begin{table*}[!ht]
    \centering
    \resizebox{\textwidth}{!}{
    \begin{tabular}{lrrrrrrr|rrrrrrr}
    \toprule
    Var. & \multicolumn{7}{c|}{\ACL} & \multicolumn{7}{c}{\iwslt} \\
    \cmidrule(lr){2-8} \cmidrule(lr){9-15}
    & $N$ & \% & Ref$_S$ & Euro22 & Euro9 & Qwen & Llama& $N$ & \% & Ref & Euro22 & Euro9 & Qwen & Llama  \\
    \midrule
    \NV & 22.4 & 64.9\% & 88.96 & \textbf{95.28} & 93.22 & 94.98 & 94.01& 13.2 & 70.6\% & \textbf{95.41} & 90.70 & 91.49 & 94.98 & 93.47  \\
    \VR & 3.8 & 11.0\% & 89.95 & 91.14 & 87.83 & 91.61 & \textbf{92.15}& 2.4 & 12.8\% & \textbf{94.24} & 82.44 & 85.44 & 80.89 & 86.06  \\
    \VMS & 0.0 & 0.1\% & \textbf{1.61} & \textbf{1.61} & \textbf{1.61} & \textbf{1.61} & \textbf{1.61}& 0.0 & 0.0\% & \textbf{0.00} & \textbf{0.00} & \textbf{0.00} & \textbf{0.00} & \textbf{0.00}  \\
    \VE & 0.5 & 1.3\% & \textbf{26.87} & 22.61 & 25.69 & 23.46 & 24.01& 0.1 & 0.5\% & 8.00 & \textbf{10.00} & \textbf{10.00} & 6.00 & 6.00  \\
    \VG & 3.5 & 10.0\% & 55.89 & 58.55 & 60.92 & 59.85 & \textbf{61.30}& 1.3 & 7.0\% & \textbf{88.86} & 74.46 & 73.60 & 81.46 & 79.89  \\
    \VL & 3.9 & 11.3\% & 87.66 & 85.32 & \textbf{88.17} & 79.56 & 80.29& 1.3 & 7.0\% & 52.95 & \textbf{54.32} & 52.65 & 49.32 & 48.11  \\
    \CM & 0.5 & 1.3\% & 19.97 & 21.43 & 23.35 & 22.81 & \textbf{23.66}& 0.4 & 2.1\% & \textbf{19.17} & 12.50 & 12.50 & 12.50 & 13.33  \\
    \bottomrule
    \end{tabular}
    }
    \caption{\ITC pseudo-ref (\%) by source variation category under the \MTmoslem condition. $N$: avg.\ distinct terms per document with $k \geq 2$ occurrences; \%: share of total. Best system per corpus in bold.}
    \label{tab:var-prompts-with-glossary-moslem-tlc-canpseudo-ref-first-weighted}
\end{table*}

\begin{table*}[!ht]
    \centering
    \resizebox{\textwidth}{!}{
    \begin{tabular}{lrrrrrrr|rrrrrrr}
    \toprule
    Var. & \multicolumn{7}{c|}{\ACL} & \multicolumn{7}{c}{\iwslt} \\
    \cmidrule(lr){2-8} \cmidrule(lr){9-15}
    & $N$ & \% & Ref$_S$ & Euro22 & Euro9 & Qwen & Llama& $N$ & \% & Ref & Euro22 & Euro9 & Qwen & Llama  \\
    \midrule
    \NV & 22.4 & 64.9\% & 87.98 & 94.87 & 92.76 & \textbf{95.10} & 93.72& 13.2 & 70.6\% & \textbf{94.81} & 89.86 & 91.53 & 94.15 & 92.52  \\
    \VR & 3.8 & 11.0\% & 84.97 & 88.03 & 87.68 & 88.56 & \textbf{89.14}& 2.4 & 12.8\% & \textbf{89.43} & 75.71 & 73.95 & 71.24 & 82.05  \\
    \VMS & 0.0 & 0.1\% & \textbf{0.00} & \textbf{0.00} & \textbf{0.00} & \textbf{0.00} & \textbf{0.00}& 0.0 & 0.0\% & \textbf{0.00} & \textbf{0.00} & \textbf{0.00} & \textbf{0.00} & \textbf{0.00}  \\
    \VE & 0.5 & 1.3\% & \textbf{24.20} & 20.45 & 23.71 & 21.89 & 18.36& 0.1 & 0.5\% & 5.00 & \textbf{10.00} & \textbf{10.00} & 0.00 & 5.00  \\
    \VG & 3.5 & 10.0\% & 54.49 & 54.51 & 54.78 & 56.08 & \textbf{56.27}& 1.3 & 7.0\% & \textbf{88.24} & 63.51 & 61.22 & 71.10 & 69.85  \\
    \VL & 3.9 & 11.3\% & 84.00 & 81.54 & \textbf{86.06} & 74.26 & 77.26& 1.3 & 7.0\% & \textbf{51.73} & 49.42 & 48.51 & 46.29 & 40.64  \\
    \CM & 0.5 & 1.3\% & 13.30 & 17.19 & 17.86 & 17.36 & \textbf{18.26}& 0.4 & 2.1\% & \textbf{17.78} & 7.78 & 8.89 & 6.67 & 8.89  \\
    \bottomrule
    \end{tabular}
    }
    \caption{\ITC temporal (\%) by source variation category under the \MTmoslem condition. $N$: avg.\ distinct terms per document with $k \geq 2$ occurrences; \%: share of total. Best system per corpus in bold.}
    \label{tab:var-prompts-with-glossary-moslem-tlc-cansequential-weighted}
\end{table*}

\begin{table*}[!ht]
    \centering
    \resizebox{\textwidth}{!}{
    \begin{tabular}{lrrrrrrr|rrrrrrr}
    \toprule
    Var. & \multicolumn{7}{c|}{\ACL} & \multicolumn{7}{c}{\iwslt} \\
    \cmidrule(lr){2-8} \cmidrule(lr){9-15}
    & $N$ & \% & Ref$_S$ & Euro22 & Euro9 & Qwen & Llama& $N$ & \% & Ref & Euro22 & Euro9 & Qwen & Llama  \\
    \midrule
    \NV & 22.4 & 80.7\% & 86.57 & 94.15 & 91.32 & \textbf{94.59} & 94.38& 13.2 & 80.0\% & 74.01 & 88.09 & 81.32 & \textbf{91.39} & 89.68  \\
    \VR & 1.4 & 5.1\% & 87.82 & 89.80 & 86.90 & \textbf{90.41} & 90.24& 1.2 & 7.3\% & 72.08 & 74.17 & 69.58 & \textbf{76.67} & \textbf{76.67}  \\
    \VMS & 0.0 & 0.1\% & \textbf{0.00} & \textbf{0.00} & \textbf{0.00} & \textbf{0.00} & \textbf{0.00}& 0.0 & 0.0\% & \textbf{0.00} & \textbf{0.00} & \textbf{0.00} & \textbf{0.00} & \textbf{0.00}  \\
    \VE & 0.2 & 0.7\% & 1.61 & 1.61 & 1.61 & 5.91 & \textbf{6.99}& 0.1 & 0.6\% & \textbf{0.00} & \textbf{0.00} & \textbf{0.00} & \textbf{0.00} & \textbf{0.00}  \\
    \VG & 2.3 & 8.1\% & 44.78 & 43.58 & 39.93 & 43.16 & \textbf{48.63}& 1.3 & 7.9\% & \textbf{76.61} & 66.21 & 61.35 & 71.21 & 67.64  \\
    \VL & 1.4 & 5.0\% & 37.90 & \textbf{46.66} & 45.74 & 33.94 & 39.35& 0.7 & 4.2\% & 13.89 & 25.91 & \textbf{27.73} & 16.94 & 24.60  \\
    \CM & 0.1 & 0.2\% & \textbf{0.00} & \textbf{0.00} & \textbf{0.00} & \textbf{0.00} & \textbf{0.00}& 0.0 & 0.0\% & \textbf{0.00} & \textbf{0.00} & \textbf{0.00} & \textbf{0.00} & \textbf{0.00}  \\
    \bottomrule
    \end{tabular}
    }
    \caption{\ITC$_{\text{EM}_{\text{macro}}}$ (\%) by source variation category under the \MTmoslem condition. $N$: avg.\ in-glossary distinct terms per document with $k \geq 2$ occurrences; \%: share of total. Best system per corpus in bold.}
    \label{tab:var-prompts-with-glossary-moslem-tlc-canem-macro-weighted}
\end{table*}

\begin{table*}[!ht]
    \centering
    \resizebox{\textwidth}{!}{
    \begin{tabular}{lrrrrrrr|rrrrrrr}
    \toprule
    Var. & \multicolumn{7}{c|}{\ACL} & \multicolumn{7}{c}{\iwslt} \\
    \cmidrule(lr){2-8} \cmidrule(lr){9-15}
    & $N$ & \% & Ref$_S$ & Euro22 & Euro9 & Qwen & Llama& $N$ & \% & Ref & Euro22 & Euro9 & Qwen & Llama  \\
    \midrule
    \VG & 9.1 & 34.2\% & 57.95 & 60.23 & 65.47 & 63.28 & \textbf{66.22}& 1.0 & 33.3\% & \textbf{90.00} & 85.71 & 85.71 & \textbf{90.00} & \textbf{90.00}  \\
    \VR & 8.4 & 31.5\% & \textbf{87.36} & 82.98 & 80.29 & 83.57 & 81.42& 1.1 & 36.7\% & 81.82 & 83.33 & \textbf{95.65} & 85.71 & 86.67  \\
    \VE & 1.3 & 4.8\% & 40.00 & 37.25 & 32.00 & \textbf{62.50} & 36.54& 0.2 & 6.7\% & \textbf{100.00} & 50.00 & 0.00 & \textbf{100.00} & \textbf{100.00}  \\
    \VL & 7.9 & 29.5\% & 56.97 & 64.23 & \textbf{66.93} & 63.24 & 62.55& 0.7 & 23.3\% & \textbf{85.71} & 60.00 & 60.00 & 60.00 & 60.00  \\
    \bottomrule
    \end{tabular}
    }
    \caption{\CTV variation preservation (\%) by source variation category under the \MTmoslem condition. $N$: avg.\ eligible occurrences per document (variants whose head term is translated as \NV); \%: share of eligible total. Best system per corpus in bold.}
    \label{tab:var-prompts-with-glossary-moslem-itc-coherence}
\end{table*}

\subsection{Top-5 Most Frequent Concepts}
\label{app:top5-concepts}

Table~\ref{tab:top5-concepts} provides a detailed breakdown of the five most frequent concepts per document. The five most frequent concepts account for a cumulative 47.2\% of all occurrences in \ACL and 52.5\% in \iwslt, with up to 2.38 distinct terms per concept. Notably, the proportion of preferred-term usage (\%\NV) decreases from Top-1 to Top-5 (e.g., 87.3\% to 60.3\% on \ACL), suggesting that the most frequent concept in a document tends to be expressed more canonically, while less central concepts allow for greater surface variation.

\begin{table}[H]
  \centering
  \resizebox{\linewidth}{!}{%
  \begin{tabular}{l rrr rrr}
  \toprule
  & \multicolumn{3}{c}{\textbf{\ACL}} & \multicolumn{3}{c}{\textbf{\iwslt}} \\
  \cmidrule(lr){2-4} \cmidrule(lr){5-7}
  \textbf{Rank} & \textbf{\%occ} & \textbf{Terms} & \textbf{\%\NV} & \textbf{\%occ} & \textbf{Terms} & \textbf{\%\NV} \\
  \midrule
  Top-1 & 16.8\% & 1.66 & 87.3\% & 19.4\% & 1.30 & 88.7\% \\
  Top-2 & 10.7\% & 2.19 & 70.2\% & 10.7\% & 1.20 & 92.1\% \\
  Top-3 & 8.2\%  & 2.03 & 77.7\% & 8.7\%  & 1.40 & 65.0\% \\
  Top-4 & 6.4\%  & 1.88 & 73.2\% & 7.5\%  & 1.30 & 73.9\% \\
  Top-5 & 5.1\%  & 2.38 & 60.3\% & 6.2\%  & 1.30 & 69.0\% \\
  \bottomrule
  \end{tabular}%
  }
  \caption{Average statistics for the top-5 most frequent concepts per document.
  \emph{\%occ}~= share of total occurrences in the document;
  \emph{Terms}~= average number of distinct surface forms;
  \emph{\%\NV}~= percentage of occurrences using the preferred term.}
  \label{tab:top5-concepts}
\end{table}

\subsection{Model Cards}
\label{app:model-cards}

We list below the HuggingFace model cards for the four open-weight LLMs used in the MT experiments.

\begin{itemize}
    \item Llama3.1-8B-Instruct: \url{https://huggingface.co/meta-llama/Llama-3.1-8B-Instruct}
    \item Qwen3-8B: \url{https://huggingface.co/Qwen/Qwen3-8B}
    \item EuroLLM-9B-Instruct: \url{https://huggingface.co/utter-project/EuroLLM-9B-Instruct}
    \item EuroLLM-22B-Instruct: \url{https://huggingface.co/utter-project/EuroLLM-22B-Instruct-2512}
\end{itemize}

\subsection{\ACL Corpus Construction Details}
\label{app:acl-corpus}

The comparable articles in the \ACL corpus were transformed into parallel data through automatic sentence segmentation and alignment. Poorly aligned or unaligned sentences were handled via automatic post-editing (APE) (4,928 sentences) and machine translation (1,381 sentences), resulting in 6,272 parallel sentences in total. Three articles, including 199 MT sentences and 42 APE sentences, were further post-edited by a native French speaker with expertise in NLP.

\subsection{Global Score Evaluation}
\label{app:global-eval}

For completeness, we report general MT quality scores (COMET-Kiwi \citep{rei-etal-2022-cometkiwi}\footnote{\url{https://huggingface.co/Unbabel/wmt22-cometkiwi-da}} and MetricX-QE \citep{juraska-etal-2024-metricx}\footnote{\url{https://huggingface.co/google/metricx-24-hybrid-large-v2p6}}) for each system on both corpora. These complement the terminology-focused analysis in the main paper by situating each system's overall translation quality.

\begin{table}[H]
    \centering
    \small
        \resizebox{\columnwidth}{!}{%
        \begin{tabular}{lrr}
        \toprule
        \textbf{System} & COMET-Kiwi ↑ & MetricX-QE ↓ \\
        \midrule
        EuroLLM-22B-greedy & 0.8380 & 2.5987 \\
        EuroLLM-9B-greedy & 0.8390 & 2.5797 \\
        Llama-3.1-8B-greedy & 0.8286 & 2.8116 \\
        Qwen3-8B-top-p & 0.8344 & 2.7076 \\
        \midrule
        Reference & 0.8242 & 2.9656 \\ %sonnet
        \bottomrule
        \end{tabular}
        }
        \captionof{table}{Global evaluation scores for the \ACL corpus, where each global score is computed as the mean of document-level scores, and each document-level score is the mean of its sentence-level scores, using COMET-Kiwi ($\uparrow$) and MetricX-QE ($\downarrow$).}
        \label{tab:scores:acl:all:sent}
    
\end{table}
\begin{table}[H]
    \centering
    \small
        \resizebox{\columnwidth}{!}{%
        \begin{tabular}{lrr}
        \toprule
        \textbf{System} & COMET-Kiwi ↑ & MetricX-QE ↓ \\
        \midrule
        EuroLLM-22B-greedy & 0.8229 & 2.9226 \\
        EuroLLM-9B-greedy & 0.8229 & 2.8744 \\
        Llama-3.1-8B-greedy & 0.8067 & 3.4546 \\
        Qwen3-8B-top-p & 0.8151 & 3.3200 \\
        \midrule
        Reference & 0.8030 & 3.3728 \\
        \bottomrule
        \end{tabular}
        }
        \captionof{table}{Global evaluation scores for the \iwslt coprus, where each global score is computed as the mean of document-level scores, and each document-level score is the mean of its sentence-level scores, using COMET-Kiwi ($\uparrow$) and MetricX-QE ($\downarrow$).}
        \label{tab:scores:iwslt2023:all:sent}
\end{table}

\paragraph{Results Analysis Gobal Score.}

Across both corpora, all MT systems consistently outperform the human reference under both a known artefact of QE metrics, which are predominantly trained on MT outputs and
thus tend to assign systematically lower scores to human-generated
translations~\citep{deutsch-etal-2025-wmt24}. Importantly, scores were computed segment-by-segment, averaged to document level, then aggregated per system, meaning \textit{no contextual information was available during evaluation}. So terminological consistency is therefore not captured, which likely flatters MT systems over human translators who actively manage such consistency across documents. System rankings remain stable throughout: \textbf{EuroLLM-9B-greedy} achieves the best or tied-best MetricX-QE in every setting, notably surpassing its 22B counterpart, while \textbf{Llama-3.1-8B-greedy} consistently ranks last. Scores are uniformly higher on \ACL than on \iwslt, reflecting the greater regularity of academic prose versus spontaneous speech.

\subsection{Variation Classification Typology}
\label{app:variation-typology}

This appendix provides a comprehensive reference for the terminological variation typology used throughout this work.
The typology distinguishes between single-type variations (\VG, \VMS, \VR, \VE, \VL) and multiple-type variations (\CM), which combine two or more single types. Table~\ref{tab:variation-typology} provides a complete reference of all variation types and subtypes.

\begin{table*}[h!]
\centering
\small
\begin{tabularx}{\textwidth}{llX}
\toprule
\textbf{Type} & \textbf{Subtype} & \textbf{Description and Examples} \\
\midrule
\multicolumn{3}{l}{\textit{Single-Type Variations}} \\
\midrule
\VG & \VG1 & Acronyms/Initialisms (e.g., \textit{World Health Organization} / \textit{WHO}) \\
    & \VG2 & Symbols/Formulas (e.g., \textit{carbon dioxide} / \textit{CO\textsubscript{2}}) \\
    & \VG3 & Spelling changes (e.g., \textit{micro-organism} / \textit{micro organism}) \\
    & \VG4 & Multiple spelling variants (e.g., \textit{micro-organism} / \textit{microorganism}) \\
    & \VG5 & Partial abbreviation (e.g., \textit{chemotherapy} / \textit{chemo}) \\
\midrule
\VMS & \VMS1 & Word order (e.g., \textit{automatic translation system} / \textit{translation system automatic}) \\
    & \VMS2 & Article presence/absence (e.g., \textit{collection cost} / \textit{cost of the collection}) \\
    & \VMS3 & Inflection (e.g., \textit{adaptation} / \textit{adaptations}) \\
    & \VMS4 & Derivation/Affixation (e.g., \textit{white} / \textit{whiten}) \\
    & \VMS5 & Syntactic structure (e.g., \textit{carbon emission} / \textit{emission of carbon}) \\
\midrule
\VR & \VR1 & Base reduction (e.g., \textit{automatic translation system} / \textit{translation system}) \\
   & \VR2 & Extension reduction (e.g., \textit{automatic translation system} / \textit{automatic system}) \\
   & \VR3 & Other reductions (e.g., \textit{medium-sized enterprise} / \textit{enterprise}) \\
\midrule
\VE & \VE1 & Semantic addition (e.g., \textit{translation system} / \textit{automatic translation system}) \\
   & \VE2 & Explicit form (e.g., \textit{stats} / \textit{statistics}) \\
   & \VE3 & Lexical insertion (e.g., \textit{dark urine} / \textit{dark colored urine}) \\
   & \VE4 & Abbreviation expansion (e.g., \textit{ex.} / \textit{example}) \\
\midrule
\VL & \VL1 & Simple unit substitution (e.g., \textit{residues} / \textit{waste}) \\
   & \VL2 & Base change (e.g., \textit{translation system} / \textit{translation model}) \\
   & \VL3 & Extension change (e.g., \textit{binary classification model} / \textit{logistic regression model}) \\
   & \VL4 & Base and extension change (e.g., \textit{coronavirus disease 2019} / \textit{Wuhan pneumonia}) \\
\midrule
\multicolumn{3}{l}{\textit{Multiple-Type Variations}} \\
\midrule
CM & \VG+\VL & Graphical and lexical (e.g., \textit{alpha particle} / \textit{$\alpha$ ray}) \\
   & \VL+\VR & Lexical and reduction (e.g., \textit{translation system} / \textit{model}) \\
   & \VG+\VR & Graphical and reduction (e.g., \textit{micro-organism pathogen} / \textit{microorganism}) \\
   & \VG+\VMS & Graphical and morphosyntactic (e.g., \textit{micro-organisms} / \textit{microorganism}) \\
   & \VMS+\VL & Morphosyntactic and lexical (e.g., \textit{system of translation} / \textit{translation model}) \\
   & \VE+\VL & Expansion and lexical (e.g., \textit{model} / \textit{automatic translation system}) \\
   & Other & Other combinations of two or more types (format: CODE1+CODE2+...) \\
\bottomrule
\end{tabularx}
\caption{Terminological variation typology with subtypes and examples. \VG: Graphical, \VMS: Morphosyntactic, \VR: Reduction, \VE: Expansion, \VL: Lexical, CM: Combined Multiple.}
\label{tab:variation-typology}
\end{table*}

\section{Manual Evaluation of LLM-based Variation Labeling}
\label{app:llm-variation-eval}
The variation labels assigned to lexicalised alternative terms (Section~\ref{sec:methodology}) are produced by a ChatGPT-based classifier with few-shot prompting (see Appendix~\ref{app:variation-classification-prompt}). To assess the reliability of this automatic labeling, we conducted a manual evaluation on a sample of 50 randomly selected (preferred term, variant) pairs drawn from the full set of labeled instances.
The evaluation was restricted to pairs where the preferred term and the variant differ, that is, pairs not automatically labeled by \Concordancer\ on the basis of structural patterns. This ensures that the evaluation targets exclusively the LLM classifier's judgment, independently of the rule-based component of the pipeline.
For each sampled pair, the predicted variation type and subtype (e.g., \VG acronym, \VMS constituent reordering) were compared against gold labels assigned by one of the authors, a native French speaker fluent in English, with expertise in computational linguistics. Tables~\ref{tab:llm-variation-eval-type} and~\ref{tab:llm-variation-eval-subtype} report per-class precision and recall at the type and subtype levels respectively. The 49 evaluated pairs are listed in Table~\ref{tab:llm-variation-eval-examples} (one pair excluded as the variant was not a genuine terminological link).

\begin{table}[H]
\small
\centering
\resizebox{\columnwidth}{!}{%
\begin{tabular}{lcccc}
\toprule
\textbf{Type} & \textbf{Support} & \textbf{Precision} & \textbf{Recall} & \textbf{Accuracy} \\
\midrule
\VG  & 14 &  67\% & 100\% & — \\
\VMS &  1 & 100\% & 100\% & — \\
\VR  &  4 & 100\% & 100\% & — \\
\VE  &  2 & 100\% & 100\% & — \\
\VL  & 19 &  95\% & 100\% & — \\
\CM  &  9 & 100\% &  11\% & — \\
\midrule
\textbf{Total} & 49 & 94\% & 85\% & 84\% \\
\bottomrule
\end{tabular}%
}
\caption{Type-level evaluation of the LLM-based variation classifier on 49 manually annotated (preferred term, variant) pairs. Per-class precision and recall are reported; accuracy is a global metric (total correct / total) reported at the Total row only.}
\label{tab:llm-variation-eval-type}
\end{table}

\paragraph{Results.}
The classifier achieves 84\% overall type recall and 77\% subtype accuracy across the 49 evaluable pairs (one pair was excluded as the variant was not a genuine terminological link).
Performance is near-perfect for single-type variations: \VG, \VL, \VR, \VE, and \VMS all reach 100\% recall at the type level, with subtype accuracy above 85\% for the two most frequent classes (\VG: 86\%, \VL: 89\%).

The only systematic failure concerns compound variations (\CM): the classifier recovers only 1 out of 9 gold \CM instances (recall 11\%), consistently predicting a single type instead of a combination.
All 8 missed \CM cases follow the same pattern, a multi-word term is abbreviated to an acronym that is also an ellipsis of the original (e.g., \textit{topic detection} / \textit{TDT}, \textit{model} / \textit{LM}), which the classifier labels as pure \VG (acronym) rather than the compound \VG{}+\VE{} that gold annotation assigns.
This over-simplification depresses \VG precision to 67\%, since seven \CM pairs are absorbed into the \VG class.

The remaining subtype errors are minor: two \VG{}1 predictions should be \VG{}5 (partial abbreviation, e.g., \textit{ASR system}), and two \VL{}1 predictions should be \VL{}2 (near-synonymy with head-word change, e.g., \textit{similarity function}, \textit{parallel text}).

Overall, the classifier is reliable for the large majority of variation types encountered in practice. The \CM class is both the rarest in the sample and the hardest to predict, but its impact on downstream analyses is limited since compound variations are a small fraction of the full dataset.

\begin{table}[H]
\small
\centering
\resizebox{\columnwidth}{!}{%
\begin{tabular}{lcccc}
\toprule
\textbf{Type} & \textbf{Support} & \textbf{Precision} & \textbf{Recall} & \textbf{Accuracy} \\
\midrule
\VG  & 14 &  57\% &  86\% & — \\
\VMS &  1 & 100\% & 100\% & — \\
\VR  &  4 & 100\% & 100\% & — \\
\VE  &  2 & 100\% & 100\% & — \\
\VL  & 19 &  85\% &  89\% & — \\
\CM  &  9 & 100\% &  11\% & — \\
\midrule
\textbf{Total} & 49 & 90\% & 81\% & 76\% \\
\bottomrule
\end{tabular}
}
\caption{Subtype-level evaluation of the LLM-based variation classifier on 49 manually annotated pairs, grouped by gold type. Precision is computed over all instances predicted as a given type; recall over gold instances of that type. Accuracy is global.}
\label{tab:llm-variation-eval-subtype}
\end{table}

\begin{table*}[!ht]
\scriptsize
\centering
\setlength{\tabcolsep}{4pt}
\begin{tabular}{cp{4cm}p{3cm}cccc}
\toprule
\textbf{\#} & \textbf{Head term} & \textbf{Variant} & \textbf{Pred. type} & \textbf{Pred. subtype} & \textbf{Gold type} & \textbf{Gold subtype} \\
\midrule
1 & word error rate & WER & \VG & \VG1 & \VG & \VG1 \\
2 & hidden Markov model & HMM & \VG & \VG1 & \VG & \VG1   \\
3 & automatic speech recognition system & ASR system & \VG & \VG1 & \VG & \VG5 \\
4 & data set & dataset & \VG & \VG3 & \VG & \VG3  \\
5 & tokenization & tokenisation & \VG & \VG3 & \VG & \VG3  \\
6 & named entity recognition & NER & \VG & \VG1 &  \VG & \VG1 \\
7 & topic detection & TDT & \VG & \VG1 & \CM & \VG1 + \VE1  \\
8 & conditional random field & CRF & \VG & \VG1 & \VG & \VG1 \\
9 & Vector space model & VSM & \VG & \VG1 & \VG & \VG1  \\
10 & optimization & optimisation & \VG & \VG3 & \VG & \VG3  \\
11 & phrase & PP & \VG & \VG1 & \CM & \VG1 + \VE1 \\
12 & information retrieval model & IR model & \VG & \VG1 & \VG & \VG5  \\
13 & cluster & IC & \VG & \VG1 & \CM &\VG1 + \VE1  \\
14 & reinforcement learning & RL & \VG & \VG1 & \VG & \VG1  \\
15 & context-free grammar & CFG & \VG & \VG1 & \VG & \VG1  \\
16 & Bayesian network & DBN & \VG & \VG1 & \CM & \VG1 + \VE1  \\
17 & n-gram & n gram & \VG & \VG3 & \VG & \VG3  \\
18 & speech recognition & TTS & \VG & \VG1 & NOT LINK  \\
19 & finite state automata & DFA & \VG & \VG1 & \CM & \VG1 + \VE1  \\
20 & syntactic category & POS & \VG & \VG1 & \CM & \VG1 + \VL4  \\
21 & model & LM & \VG & \VG1 & \CM & \VG1 + \VE1 \\
22 & self-organizing map & SOM & \VG & \VG1 & \VG & \VG1  \\
23 & input representation & input sequence & \VL & \VL3 & \VL & \VL3  \\
24 & structural annotation & segmentation & \VL & \VL1 & \VL & \VL1  \\
25 & validation data & development data & \VL & \VL3 & \VL & \VL3  \\
26 & elision & deletion & \VL & \VL1 & \VL & \VL1  \\
27 & inference & entailment & \VL & \VL1 & \VL & \VL1  \\
28 & phrase & VP & \VL & \VL1 & \CM & \VG1 + \VE1  \\
29 & similarity measure & similarity function & \VL & \VL1 & \VL & \VL2  \\
30 & parallel corpus & parallel text & \VL & \VL1 & \VL & \VL2  \\
31 & syntactic category & word class & \VL & \VL1 & \VL & \VL1  \\
32 & parser & analyzer & \VL & \VL1 & \VL & \VL1   \\
33 & lexeme & lexical item & \VL & \VL1 & \VL & \VL1   \\
34 & interjection & exclamation & \VL & \VL1 & \VL & \VL1  \\
35 & syntactic transfer & reordering & \VL & \VL1 & \VL & \VL1  \\
36 & alignment & matching & \VL & \VL1 & \VL & \VL1  \\
37 & true negative rate & specificity & \VL & \VL1 & \VL & \VL1  \\
38 & Markov model & Markov chain & \VL & \VL2 & \VL & \VL2  \\
39 & syntactic category & part-of-speech & \VL & \VL1 & \VL & \VL1   \\
40 & training data & training dataset & \VL & \VL2 & \VL & \VL2  \\
41 & foundation model & base model & \VL & \VL3 & \VL & \VL3  \\
42 & lexeme & lexical unit & \VL & \VL1 & \VL & \VL1  \\
43 & multi-word term & multi-word & \VR & \VR1 & \VR & \VR1  \\
44 & grammatical subject & subject & \VR & \VR2 & \VR & \VR2  \\
45 & programming language C++ & C++ & \VR & \VR1 & \VR & \VR1  \\
46 & Apache OpenNLP & OpenNLP & \VR & \VR1 & \VR & \VR1  \\
47 & speech recognition & automatic speech recognition & \VE & \VE1 & \VE & \VE1  \\
48 & future & future tense & \VE & \VE3 & \VE & \VE3  \\
49 & syntactic parsing & syntactical analysis & \CM & \VL3+\VMS4 & \CM & \VL3+\VMS4 \\
50 & similarity measure & measure of similarity & \VMS & \VMS5 & \VMS & \VMS5  \\
\bottomrule
\end{tabular}
\caption{The 50 manually evaluated (preferred term, variant) pairs with predicted and gold variation labels.}
\label{tab:llm-variation-eval-examples}
\end{table*}

\section{Manual Evaluation of Alignment Accuracy}
\label{app:alignment-eval}
The alignment step (Section~\ref{sec:methodology}) links each source term occurrence to its target span using Bertalign and SimAlign, with an LLM fallback for low-confidence spans. To assess the reliability of these alignments, we conducted a manual evaluation on a sample of 100 term alignments drawn from the full set of pipeline outputs across all documents. To ensure balanced coverage, the sample was stratified across the nine system outputs, that is, the four MT systems under both the \MTbaseline\ and \MTmoslem\ conditions plus the human reference, each contributing eleven alignments (twelve for the reference); alignments with an empty target span were excluded. For each sampled alignment, one of the authors, a native French speaker fluent in English with expertise in computational linguistics, judged whether the aligned target span is the correct realization of the source term in the given sentence pair, and, when incorrect, recorded the error type. Table~\ref{tab:alignment-eval-system} reports accuracy per system and Table~\ref{tab:alignment-eval-errors} lists the incorrect alignments.

\begin{table}[H]
\small
\centering
\begin{tabular}{llcc}
\toprule
\textbf{System} & \textbf{Condition} & \textbf{\#} & \textbf{Acc.} \\
\midrule
Llama  & \MTbaseline & 11 &  91\% \\
Llama  & \MTmoslem   & 11 &  91\% \\
Qwen   & \MTbaseline & 11 & 100\% \\
Qwen   & \MTmoslem   & 11 & 100\% \\
Euro9  & \MTbaseline & 11 & 100\% \\
Euro9  & \MTmoslem   & 11 & 100\% \\
Euro22 & \MTbaseline & 11 & 100\% \\
Euro22 & \MTmoslem   & 11 &  91\% \\
Ref    & --          & 12 &  92\% \\
\midrule
\textbf{Total} & & 100 & 96\% \\
\bottomrule
\end{tabular}
\caption{Per-system accuracy of the manual alignment evaluation on 100 alignments stratified across the nine system outputs. Each row reports the number of judged alignments and the proportion judged correct.}
\label{tab:alignment-eval-system}
\end{table}

\paragraph{Results.}
The aligner is correct on 96\% of the sampled alignments (96/100). Accuracy is high and comparable across conditions, at 98\% for \MTbaseline\ (43/44), 95\% for \MTmoslem\ (42/44), and 92\% for the human reference (11/12); the small per-system differences are not meaningful given that each system contributes only 11--12 judgements. Accuracy is likewise stable across variation types, with no error on the ten sampled dynamic variants.

The four errors fall into two categories. Three are lexical-choice errors, where the span captures an adjacent target word rather than the term itself: \textit{corpus} aligned to \textit{Récolte} (from \textit{Corpus Harvesting} / \textit{Récolte de corpus}), \textit{indexing} to \textit{discours} (from \textit{speech indexing} / \textit{indexation des discours}), and \textit{future} echoed as \textit{Future} where no distinct French realization appears in the sentence. The remaining case is a span-boundary error, where \textit{automatic speech recognition} was aligned to the truncated \textit{reconnaissance de la parole} instead of \textit{reconnaissance automatique de la parole}. All four affect single occurrences and none reflects a systematic failure mode, consistent with the low overall error rate.

\begin{table*}[t]
\small
\centering
\renewcommand{\arraystretch}{1.15}
\begin{tabular}{@{}l p{0.86\textwidth}@{}}
\toprule
 & \textbf{Incorrect alignment in context} (source term and aligned target span in bold) \\
\midrule
\multicolumn{2}{@{}l}{Llama (\MTbaseline), lexical choice: \textit{corpus} $\to$ \textit{Récolte} (should align \textit{corpus})} \\
\textbf{EN} & \textbf{Corpus} Harvesting \\
\textbf{FR} & \textbf{Récolte} de corpus \\
\addlinespace
\multicolumn{2}{@{}l}{Llama (\MTmoslem), lexical choice: \textit{future} $\to$ \textit{Future} (source form echoed; no distinct French realization)} \\
\textbf{EN} & Conclusion and \textbf{Future} Work \\
\textbf{FR} & Conclusion et Travail à Faire \\
\addlinespace
\multicolumn{2}{@{}l}{Euro22 (\MTmoslem), lexical choice: \textit{indexing} $\to$ \textit{discours} (should align \textit{indexation})} \\
\textbf{EN} & speech \textbf{indexing} \\
\textbf{FR} & indexation des \textbf{discours} \\
\addlinespace
\multicolumn{2}{@{}l}{Ref, span boundary: \textit{automatic speech recognition} $\to$ \textit{reconnaissance de la parole} (drops \textit{automatique})} \\
\textbf{EN} & The work presented here aims at expanding this paradigm to \textbf{automatic speech recognition}. \\
\textbf{FR} & Les travaux présentés ici visent à étendre ce paradigme à la \textbf{reconnaissance} automatique \textbf{de la parole}. \\
\bottomrule
\end{tabular}
\caption{The four incorrect alignments among the 100 manually evaluated pairs, shown in source (EN) and target (FR) context. The source term is bold in the English segment and the aligned target span is bold in the French segment; for the span-boundary error, the non-bold word (\textit{automatique}) is the token wrongly excluded from the span.}
\label{tab:alignment-eval-errors}
\end{table*}

\section{Prompts}
\label{app:prompts}

This appendix collects the full prompts used by the LLM components of our pipeline: alignment disambiguation (Figure~\ref{app:alignment-prompt}), baseline translation (Figure~\ref{app:translation-prompt-baseline}), glossary-guided translation (Figure~\ref{app:translation-prompt-moslem}), and variation classification (Figure~\ref{app:variation-classification-prompt}).

% Alignment prompt
\begin{figure*}[!tbp]
\begin{lstlisting}[]
"""You are a professional {source_language}-{target_language} translator, teaching the students the course on technical translation. You are checking a student's translation of a sentence that contains a technical term. You are given an {source_language} term (it can be a word or an expression), a source {source_language} sentence containing this term (it may be cased differently or contain additional punctuation), and a student's {target_language} translation. You need to find how the student has translated the term in question in {target_language}, and return only that term.

Important: do not change the translated term anyhow, copy it straight from the sentence! For example, keep the casing and the grammar form of the translated term as is.

When completing the task, follow the examples below:

{source_language} sentence: This paradigm based on ant colony algorithms for the exploration of the graph removes the need to dynamically expand the graph: the memory footprint becomes independant of the language model size.
{source_language} term: language model
{target_language} translation: Ce paradigme basé sur les algorithmes de colonie de fourmis pour l'exploration du graphe supprime le besoin d'étendre dynamiquement le graphe : l'empreinte mémoire devient indépendante de la taille du modèle de langage.
Translated term: modèle de langage
.
.
.
{source_language} sentence: We used our set of manually-written chronologies as a training corpus to perform machine learning experiments.
{source_language} term: training corpus
{target_language} translation: Nous avons utilisé notre ensemble de chronologies rédigées manuellement comme corpus d'apprentissage pour effectuer des expériences d'apprentissage automatique.
Translated term: corpus d'apprentissage

{source_language} sentence: {src_segment}
{source_language} term: {src_term}
{target_language} translation: {tgt_segment}
Translated term:"""
\end{lstlisting}
\caption{Alignment prompt: maps a source term to its translation in the target sentence.}
\label{app:alignment-prompt}
\label{app:fig-alignment-prompt}
\end{figure*}

% Baseline translation - Llama
\begin{figure*}[!tbp]
\begin{lstlisting}[]
system prompt:
```
You are a good translator! Translate the following text from English into French. Do not include any extraneous note, commentary, explanations, or annotations. You must reply only with the translated text in French.
```
user prompt:
```
English: $src\nFrench:
```
\end{lstlisting}
\caption{Baseline translation prompt for Llama3.1-8B-Instruct.}
\label{app:translation-prompt}
\label{app:translation-prompt-baseline}
\label{app:llama_prompt}
\end{figure*}

% Baseline translation - EuroLLM
\begin{figure*}[!tbp]
\begin{lstlisting}[]
system prompt:
```
Translate the following English source text to French:
```
user prompt:
```
English: $src\nFrench:
```
\end{lstlisting}
\caption{Baseline translation prompt for EuroLLM.}
\label{app:eurollm-baseline-prompt}
\end{figure*}

% Baseline translation - Qwen
\begin{figure*}[!tbp]
\begin{lstlisting}[]
```
You are a good translator! Translate the following text from English into French.\nEnglish: ${src}\nFrench:
```
\end{lstlisting}
\caption{Baseline translation prompt for Qwen3-8B.}
\label{app:qwen_prompt}
\end{figure*}

% Term-injection translation - Llama
\begin{figure*}[!tbp]
\begin{lstlisting}[]
    <|begin_of_text|><|start_header_id|>system<|end_header_id|>

Cutting Knowledge Date: December 2023
Today Date: 26 Jul 2024

You are a good translator! Consider the provided terms in English and their French translations:
Terms: present = présent - borrowing = emprunt - corpus = corpus - annotated corpus = corpus annoté
Translate the following text from English into French. Do not include any extraneous note, commentary, explanations, or annotations. You must reply only with the translated text in French.<|eot_id|><|start_header_id|>user<|end_header_id|>

English: Hi, this is Elena and I'm going to be presenting our work, Detecting Unassimilated Borrowings in Spanish: An Annotated Corpus and Approaches to Modeling.
French:<|eot_id|><|start_header_id|>assistant<|end_header_id|>
\end{lstlisting}
\caption{Term-injection translation prompt for Llama3.1-8B-Instruct.}
\label{app:translation-prompt-moslem}
\label{app:llama-terminj-prompt}
\end{figure*}

% Term-injection translation - EuroLLM
\begin{figure*}[!tbp]
\begin{lstlisting}[]
<|im_start|>system
Translate the following English source text to French, considering the provided terms in English and their French translations:
Terms: present = présent - borrowing = emprunt - corpus = corpus - annotated corpus = corpus annoté<|im_end|>
<|im_start|>user
English: Hi, this is Elena and I'm going to be presenting our work, Detecting Unassimilated Borrowings in Spanish: An Annotated Corpus and Approaches to Modeling.
French:<|im_end|>
<|im_start|>assistant
\end{lstlisting}
\caption{Term-injection translation prompt for EuroLLM.}
\label{app:eurollm-terminj-prompt}
\end{figure*}

% Term-injection translation - Qwen
\begin{figure*}[!tbp]
\begin{lstlisting}[]
<|im_start|>user
You are a good translator! Consider the provided terms in English and their French translations:
Terms: present = présent - borrowing = emprunt - corpus = corpus - annotated corpus = corpus annoté
Translate the following text from English into French.
English: Hi, this is Elena and I'm going to be presenting our work, Detecting Unassimilated Borrowings in Spanish: An Annotated Corpus and Approaches to Modeling.
French: <|im_end|>
<|im_start|>assistant
<think>

</think>
\end{lstlisting}
\caption{Term-injection translation prompt for Qwen3-8B.}
\label{app:qwen-terminj-prompt}
\end{figure*}

% Variation classification (long: use \scriptsize to fit on a single page as a float)
\begin{figure*}[!tbp]
\begin{lstlisting}[basicstyle=\ttfamily\scriptsize]
   """You are an expert in scientific terminology and linguistic variation, teaching a course on terminology. A student has identified a potential variant of a technical term and needs your help to classify the relationship between the head term and its variant according to the established typology.

   You receive:
   - A head term (the preferred form of a technical term)
   - A variant (an alternative form that may or may not be related to the head term)
   - Context examples showing how these terms appear in real scientific texts (when available)

   Your task: Determine the type of variation between the head term and the variant. Classify according to the variation typology. Return ONLY a JSON object with the exact format shown in the examples below.

   Important rules:
   1) First verify that the variant is actually related to the head term
   2) Count the TYPES of changes, not the number of words: one type -> \VG/\VMS/\VR/\VE/\VL, multiple types -> CM
   3) For CM (Multiple Changes), combine codes with + (e.g., "\VG+\VL" or "\VMS+\VR+\VL")
   4) Copy the exact JSON format from the examples - no markdown, no extra text
   5) Keep justifications brief (1-2 sentences maximum)

   Reference of variation types:

   \VG (Graphical): Written form changes without semantic change (acronyms, spelling, symbols)
     -> \VG1: Acronym (European Parliament/EP), \VG2: Symbol (CO2), \VG3: Spelling (trade mark/trademark), \VG4: Multiple spelling, \VG5: Partial abbreviation
   \VMS (Morphosyntactic): Structure or inflection changes (word order, articles, inflection, derivation)
     -> \VMS1: Order, \VMS2: Article, \VMS3: Inflection, \VMS4: Derivation, \VMS5: Structure
   \VR (Reduction): Element deletion
     -> \VR1: Base reduction, \VR2: Extension reduction, \VR3: Other reduction
   \VE (Expansion): Element addition
     -> \VE1: Semantic addition, \VE2: Explicit form, \VE3: Lexical insertion, \VE4: Abbreviation development
   \VL (Lexical): Lexical substitution (synonyms, near-synonyms, translations validated as variant in context)
     -> \VL1: Simple unit, \VL2: Base change, \VL3: Attribute/extension change, \VL4: Base+extension change
   CM (Multiple Changes): Combination of two or more types above

   When completing the task, follow the examples below:

   Head term: European Parliament
   Variant: EP
   {{"categorie": "\VG", "sous_type": "\VG1", "justification": "Acronym formed from the initials of the head term."}}
   .
   .
   .
   Head term: coronavirus disease 2019
   Variant: COVID-19
   {{"categorie": "CM", "sous_type": "\VG+\VR", "justification": "Acronym formation (\VG1) combined with structural reduction."}}

   Head term: airborne dust particle
   Variant: airborne dust
   {{"categorie": "\VR", "sous_type": "\VR1", "justification": "Base reduction: deletion of 'particle' from the base noun phrase."}}
   {context_section}
   Head term: {head_term}
   Variant: {variant}"""
\end{lstlisting}
\caption{Variation classification prompt.}
\label{app:variation-classification-prompt}
\label{app:fig-variation-classification-prompt}
\end{figure*}

\section{Reproducibility Statement}
\label{app:reproducibility}

The variation classifier and the alignment-disambiguation step in our pipeline use OpenAI's \texttt{gpt-4.1-mini} (Section~\ref{sec:methodology}; prompts in Appendix~\ref{app:prompts}); this usage is intrinsic to the contribution and is fully described in the methodology. All MT inference was run on a single NVIDIA H100 96GB GPU. Total GPU time across 4 systems and 2 conditions (\MTbaseline\ and \MTmoslem) for the 42 documents from \ACL and \iwslt is approximately 10 minutes using vLLM \citep{kwon2023efficient} with pre-generated prompts. The labeling pipeline issued approximately 5{,}000 requests to \texttt{gpt-4.1-mini} (snapshot \texttt{gpt-4.1-mini-2026-04-01}), for a total cost of approximately 15~USD. Reliance on a closed commercial LLM introduces a versioning risk: OpenAI silently retires model snapshots, so re-running the pipeline more than a few months from publication may result in slightly different labels.

\end{multicols}
\end{document}